\documentclass[letterpaper, 10 pt, conference]{ieeeconf}  

\IEEEoverridecommandlockouts                              

\usepackage{amsmath}
\usepackage{mathtools}
\usepackage{tabularx}
\usepackage{amsmath, amssymb}
\usepackage{hyperref, graphicx, cite, multirow}
\usepackage{subcaption}

\hypersetup{
	colorlinks=true,
	linkcolor=blue,
	citecolor=blue,
	urlcolor=blue,
}

\title{\LARGE \bf
Deep Reinforcement Learning for Autonomous Ground Vehicle Exploration Without A-Priori Maps
}

\author{Shathushan Sivashangaran and Azim Eskandarian
\thanks{\textit{Corresponding author: Shathushan Sivashangaran}}%
\thanks{The authors are with the Autonomous Systems and Intelligent Machines Laboratory, Virginia Tech, Blacksburg, VA 24061, USA.\indent(email: \href{mailto:shathushansiva@vt.edu}{shathushansiva@vt.edu}; \href{mailto:eskandarian@vt.edu}{eskandarian@vt.edu}).}
\thanks{Digital Object Identifier (DOI): 10.54364/AAIML.2023.1170}
\thanks{© 2023 the authors. This work has been accepted to Advances in Artificial Intelligence and Machine Learning for publication under a Creative Commons License CC BY 4.0.}}

\begin{document}

\maketitle
\thispagestyle{empty}
\pagestyle{empty}

\begin{abstract}

Autonomous Ground Vehicles (AGVs) are essential tools for a wide range of applications stemming from their ability to operate in hazardous environments with minimal human operator input. Effective motion planning is paramount for successful operation of AGVs. Conventional motion planning algorithms are dependent on prior knowledge of environment characteristics and offer limited utility in information poor, dynamically altering environments such as areas where emergency hazards like fire and earthquake occur, and unexplored subterranean environments such as tunnels and lava tubes on Mars. We propose a Deep Reinforcement Learning (DRL) framework for intelligent AGV exploration without a-priori maps utilizing Actor-Critic DRL algorithms to learn policies in continuous and high-dimensional action spaces directly from raw sensor data. The DRL architecture comprises feedforward neural networks for the critic and actor representations in which the actor network strategizes linear and angular velocity control actions given current state inputs, that are evaluated by the critic network which learns and estimates Q-values to maximize an accumulated reward. Three off-policy DRL algorithms, DDPG, TD3 and SAC, are trained and compared in two environments of varying complexity, and further evaluated in a third with no prior training or knowledge of map characteristics. The agent is shown to learn optimal policies at the end of each training period to chart quick, collision-free exploration trajectories, and is extensible, capable of adapting to an unknown environment without changes to network architecture or hyperparameters. The best algorithm is further evaluated in a realistic 3D environment. 

\end{abstract}

\section{INTRODUCTION}

Autonomous Ground Vehicles (AGVs) are indispensable tools for mapping uncharted terrain, Search and Rescue (SAR) missions, disaster response, military operations, mining, and extraterrestrial planetary exploration owing to their ability to operate in hazardous, unstructured environments reliably with minimal input from a human operator \cite{sun2021motion, victerpaul2017path}. Perceiving the environment, and planning trajectories are key components of AGV navigation.

Mobile robot trajectories require optimization for shortest path, minimum energy consumption and training time \cite{victerpaul2017path}. Conventional AGV navigation algorithms are dependent on specific environmental configurations \cite{levinson2011towards} which limits their effectiveness in adapting to information poor, dynamically changing environments such as areas where emergency hazards like fire and earthquake occur, and unexplored subterranean environments such as tunnels, caves and lava tubes on Mars.

Conventional navigation algorithms comprise graph search algorithms such as Dijkstra, A* and D* \cite{wang2011application} that are well-defined and simple to use but are inefficient in complex, dynamic environments and have poor robustness to noise interference and errors in the environment model, random sampling algorithms such as Probability Graph Method (PGM) and Rapid exploration Random Tree (RRT) \cite{gammell2014informed} that select random scatter points in the entire environment space to search for the optimal path between the starting and end points making them susceptible to poor real-time performance, sub-optimal solutions and high computation cost, Artificial Potential Field (APF) \cite{vadakkepat2000evolutionary} that is efficient but prone to local minima traps, and nature inspired algorithms such as fuzzy logic that is robust, but requires prior knowledge in the form of user defined knowledge based logic and rules, and Genetic Algorithm (GA) \cite{bakdi2017optimal} which is ideal for the global optimal solution and suitable for complex problems, but has poor local search ability and slow convergence rate.

Recent advancements in Artificial Intelligence (AI), sensors, communication and computer technology facilitate intelligent AGVs capable of high autonomy. Motion planning models that incorporate Artificial Neural Networks (ANNs) and Actor-Critic Reinforcement Learning (RL) enable robotic systems to learn optimal, end-to-end policies in continuous and high-dimensional action spaces directly from characteristics of high-dimensional sensory input data to intelligently select goal driven actions in obstacle filled unstructured terrain in the absence of prior knowledge and detailed maps \cite{sivashangaran2021Thesis, sivashangaran2021intelligent, zhu2021deep, larsen2021comparing}.  

Numerous works have focused on utilizing Deep Reinforcement Learning (DRL) for task-driven navigation aided by GPS observations to specific goal positions generated by a high-level planner. In \cite{wijmansdd}, the authors train an indoor robot equipped with camera, GPS and compass sensors in simulation, to navigate to a target location from a random initial position in an unseen map, for over 180 days of GPU-time parallelly with 64 GPUs, in 3 days using Decentralized Distributed Proximal Policy Optimization (DD-PPO) to achieve exceptional post-training performance. 

However, task-independent exploration of a new environment to facilitate various down-stream applications has received significantly less attention \cite{chen2019learning}. Effective exploration in GPS-denied environments such as indoor and subterranean environments encountered in SAR, military and extraterrestrial planetary exploration require even greater training times, and different DRL approaches.

Simultaneous Localization And Mapping (SLAM) enables AGVs to simultaneously estimate vehicle state utilizing on-board sensors and construct a model of the environment the sensors perceive \cite{cadena2016SLAMsurvey}. The inclusion of LIDAR and visual SLAM in the perception pipeline is a key enabler for contemporary AGV navigation in environments that are GPS-denied with no access to a-priori maps \cite{ebadi2022present}. In \cite{chaplotlearning}, the authors propose Active Neural SLAM (ANS), a modular navigation model that integrates SLAM with DRL to effectively explore an unknown environment. ANS comprises a neural SLAM module to predict the map of the surrounding environment and the agent pose, a global policy that uses a convolutional neural network to output a long-term goal using the spatial map, agent pose and visited locations as inputs trained using Proximal Policy Optimization (PPO) \cite{schulman2017proximal} to maximize coverage, and a local policy that utilizes RGB observations and a short-term goal derived from the long-term goal as inputs to a recurrent neural network to obtain the navigational action. 

SLAM is subject to mapping and pose estimate errors as a consequence of estimating sequential motion, and a modular navigation model is computationally demanding as it comprises multiple ANNs for each subtask. End-to-end navigation utilizing a single DRL network achieved with reward shaping without a dependence on mapping and long-term goal generation is of benefit to early stage task-independent exploration of unknown environments in adverse environmental conditions for data collection to build more accurate maps, and facilitate down-stream applications.   

Moreover, on-policy Actor-Critic Deep Reinforcement Learning (DRL) algorithms such as PPO, that are most commonly used in prior work, are robust to hyperparameter tuning and straightforward to implement, but are sample inefficient as these require new training samples for every policy update, which makes learning an effective policy for complex tasks computationally exorbitant. Off-policy Actor-Critic DRL algorithms such as Deep Deterministic Policy Gradient (DDPG) \cite{lillicrap2015continuous}, Twin Delayed Deep Deterministic Policy Gradient (TD3) \cite{fujimoto2018addressing} and Soft Actor-Critic (SAC) \cite{haarnoja2018soft} reuse past experience stored in a replay buffer for learning, thus have higher sample efficiency.

Given the potential of DRL for AGV navigation in information poor environments, this paper presents and evaluates a DRL framework with shaped rewards for cognitive AGV navigation solely utilizing distance measurement observations obtained using LiDAR or camera RGB-D point cloud, and compares state-of-the-art off-policy DRL algorithms' ability to safely navigate and explore obstacle filled terrain without prior knowledge of environment characteristics. Range measurements were chosen as observations over RGB images as they are more efficient, since they provide a complete understanding of the surrounding environment at a fraction of the data size. Animals such as bats use echolocation for navigation \cite{jones2007bat} solely utilizing an understanding of distances to objects, hence it is reasonable to expect a DRL agent to learn a reliable, robust navigation policy exclusively utilizing range measurements. The decrease in the size of the observation space enables faster learning of policies in more complex environments that require a large number of training samples. 

The rest of the paper is organized as follows: Section II provides a background on DRL. The proposed DRL framework is detailed in Section III. The training and evaluation methodologies are described in Section IV. The results are presented and analyzed in Section V, and finally Section VI provides closing remarks and future work directions.

\section{BACKGROUND ON DEEP REINFORCEMENT LEARNING}

\begin{figure}[thpb]
      \centering
      \includegraphics[scale=0.2]{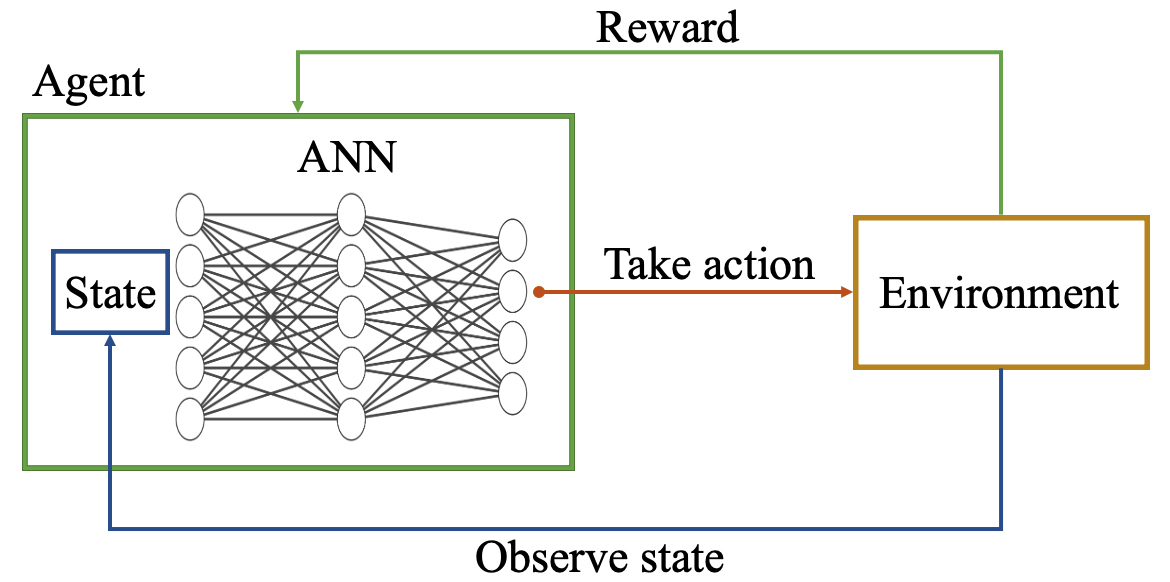}
      \caption{Schematic of deep reinforcement learning framework.}
      \label{DRL}
\end{figure}

RL is a Machine Learning (ML) framework inspired by trial-and-error animal learning to train agents that interact with the surrounding environment by promoting or discouraging actions utilizing reward feedback signals designed to gauge effectiveness of executed actions. Deep Learning (DL), a key ML component, utilizes ANNs to form an abstract, distinguishable high-level representation from low-level input features. DRL algorithms combine DL and RL to extract unknown environment features from high-dimensional input data utilizing ANNs, and decide control actions using RL. Figure \ref{DRL} portrays the DRL framework.

A RL agent observes its environment $s_{i}$ at each time step \textit{t}, and selects an action $a_{i}$ from action space \textit{A}, conforming to a learned policy $\pi(a_{i}\mid s_{i})$ that maps states to actions. The expectation of a discounted, accumulated reward $R_{i} = \Sigma_{k=0}^{\infty}\gamma^{k}rw_{i+k}$ at each state is maximized during learning, where $\gamma$ $\in$ (0,1] is the discount factor, and $rw_{i}$ is the scalar reward signal for selecting action $a_{i}$ \cite{sutton2018reinforcement}.

\subsection{Actor-Critic Framework}

\begin{figure}[!h]
    \centering
    \includegraphics[scale=0.21]{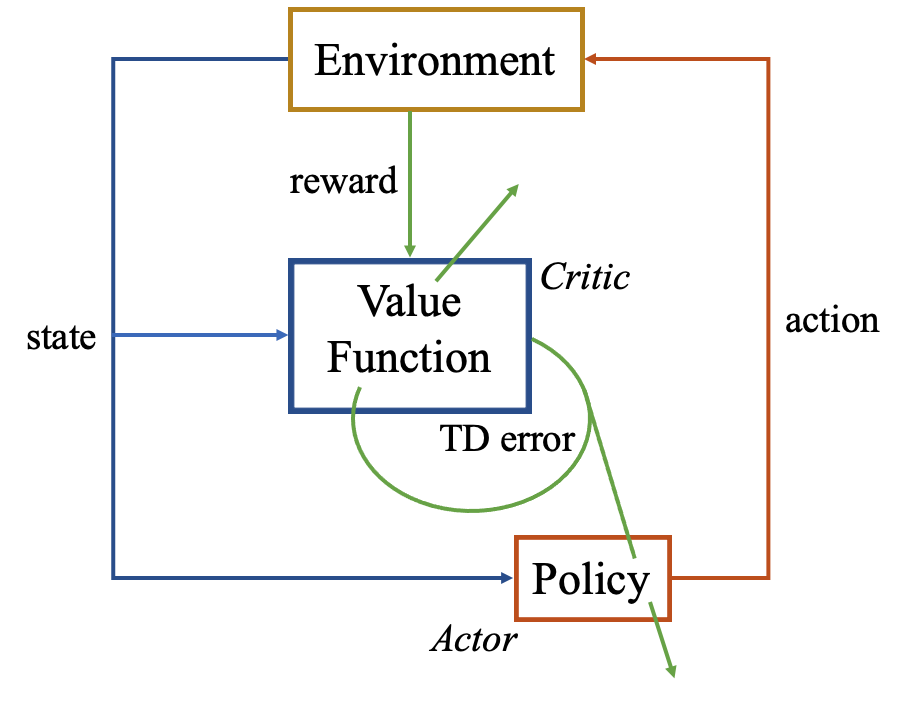}
    \setlength{\abovecaptionskip}{-\baselineskip}
    \caption{Schematic of Actor-Critic framework.} \label{actorcritic}
\end{figure}

An actor-critic framework utilizing deep function approximators that combines both value-based and policy-based RL is the preferred method to learn policies in continuous and high-dimensional action spaces, required for robotics applications. This method leverages the joint computing and decision-making abilities of the actor and critic neural networks to yield low variance and fast speeds when updating gradients. Figure \ref{actorcritic} illustrates the Actor-Critic framework.

The actor network strategizes an action output selected from a continuous action space using policy gradient, utilizing the current state as the input. The critic evaluates the chosen actions and outputs the associated approximate Q-value for the current state and selected action using an approximated value function to counter the large variance in the policy gradients. In off-policy algorithms, sample data accumulated in a replay buffer is utilized to update and approximate the value function yielding higher sample efficiency than on-policy algorithms. The two networks compute the action prediction for the current state at each time step to generate a temporal-difference error signal. 

\subsection{Deep Deterministic Policy Gradient}

DDPG \cite{lillicrap2015continuous} is a model-free, off-policy actor-critic RL algorithm that combines ANNs with the actor-critic representation of standard Deterministic Policy Gradient (DPG) \cite{silver2014deterministic} to successfully implement control sequences in a continuous action space. The actor, $\pi(s\mid\theta)$ and critic, $Q(s,a\mid\phi)$ each comprise fully-linked, two-layer feedforward ANNs with Rectified Linear Unit (ReLU) activation functions. 

The loss $L$ is minimized across all sampled experiences to update the critic parameters, $\phi$,

\begin{equation} \label{}
L = \frac{1}{M}\sum_{i=1}^{M}(y_{i} - Q(s_{i},a_{i}\mid\phi))^{2}
\end{equation}

Here $M$ is a random mini-batch of experiences, and $y_{i}$ is the target value function computed as follows,

\begin{equation} \label{}
y_{i} = R_{i} + \gamma Q_{t}(s_{i+1},\pi_{t}(s_{i+1}\mid\theta_{t})\mid\phi_{t})
\end{equation}

$\theta_{t}$ and $\phi_{t}$ are parameters of the target actor $\pi_{t}$ and target critic $Q_{t}$ respectively, that have the same structure and parameterization as $\pi$ and $Q$. The agent periodically updates $\theta_{t}$ and $\phi_{t}$ using the latest $\theta$ and $\phi$ values to improve the stability of the optimization. 

The actor parameters, $\theta$ are updated using a sampled policy gradient $\nabla_{\theta}J$ to maximize the expected discounted reward,

\begin{equation} \label{}
\nabla_{\theta}J \approx \frac{1}{M}\sum_{i=1}^{M}G_{ai}G_{\pi i}
\end{equation}

Here $G_{ai}$ is the gradient of the critic output with respect to the action selected by the actor network computed as follows,

\begin{equation} \label{}
G_{ai} = \nabla_{a}Q(s_{i},\pi(s_{i}\mid\theta)\mid\phi)
\end{equation}

$G_{\pi i}$ is the gradient of the actor output with respect to its parameters,

\begin{equation} \label{}
G_{\pi i} = \nabla_{\theta}\pi(s_{i}\mid\theta)
\end{equation}

\subsection{Twin-Delayed Deep Deterministic Policy Gradient}

TD3 is designed to improve learned policies by preventing overestimation of the value function \cite{fujimoto2018addressing}. Two Q-value functions are learned simultaneously, and the minimum is used for policy updates. Moreover, the policy is updated less frequently than the Q-value function to further improve learned policies.

The parameters of the critic, $Q_{k}(s,a\mid\phi_{k})$, where $k = 2$ is the number of critics, are updated by minimizing the loss $L_{k}$ as follows,

\begin{equation} \label{}
L_{k} = \frac{1}{M}\sum_{i=1}^{M}(y_{i} - Q_{k}(s_{i},a_{i}\mid\phi_{k}))^{2}
\end{equation}

The target value function $y_{i}$ is computed as follows,

\begin{equation} \label{}
y_{i} = R_{i} + \gamma\displaystyle\min_{k} (Q_{tk}(s_{i+1}, clip(\pi_{t}(s_{i+1}\mid\theta_{t})+\varepsilon)\mid\phi_{tk}))
\end{equation}

Here $\theta_{t}$ and $\phi_{tk}$ are parameters of the target actor $\pi_{t}$ and target critics $Q_{tk}$, and $\varepsilon$ is noise added to the computed action to promote exploration. The action is clipped based on the noise limits.

The actor parameters are updated similar to DDPG using Equation (3) where $G_{ai}$ is computed as follows and $G_{\pi i}$ is computed as in Equation (5).

\begin{equation} \label{}
G_{ai} = \nabla_{a}\displaystyle\min_{k}(Q_{k}(s_{i},\pi(s_{i}\mid\theta)\mid\phi))
\end{equation}

\subsection{Soft Actor-Critic}

SAC \cite{haarnoja2018soft}, similar to DDPG and TD3, is a model-free, off-policy actor-critic RL algorithm. In addition to maximizing the long-term expected reward, SAC maximizes the entropy of the policy, which is a measure of the policy uncertainty at a given state. A higher policy entropy promotes exploration, hence the learned policy balances exploitation and exploration of the environment.

The agent utilizes a stochastic actor that outputs mean and standard deviation, using which an unbounded action is randomly selected from a Gaussian distribution. The entropy of the policy is computed during training for the given observation using this unbounded probability distribution. Bounded actions that comply with the action space are generated by applying $tanh$ and scaling operations to the unbounded action.

The critic parameters are updated at specific time step periods by minimizing the loss function in Equation (6), similar to TD3 for $k$ critics.

The target value function $y_{i}$ is computed as the sum of the minimum discounted future reward from the critic networks $R_{i}$, and the weighted entropy as follows,

\begin{equation} \label{}
\begin{array}{c}
y_{i} = R_{i} + \gamma\displaystyle\min_{k} (Q_{tk}(s_{i+1}, \pi(s_{i+1}\mid\theta)\mid\phi_{tk}))\\
\cr - \alpha ln\pi(s_{i+1}\mid\theta) 
\end{array}
\end{equation}

Here $\alpha$ is the entropy loss weight. The entropy weight is updated by minimizing the loss function, $L_{\alpha}$ where $H$ is the target entropy as follows,

\begin{equation} \label{}
L_{\alpha} = \frac{1}{M}\sum_{i=1}^{M}(- \alpha ln\pi(s_{i}\mid\theta) - \alpha H)
\end{equation}

The stochastic actor parameters are updated by minimizing the objective function $J_{\pi}$, 

\begin{equation} \label{}
J_{\pi} = \frac{1}{M}\sum_{i=1}^{M}(-\displaystyle\min_{k} (Q_{tk}(s_{i}, \pi(s_{i}\mid\theta)\mid\phi_{tk})) + \alpha ln\pi(s_{i}\mid\theta))
\end{equation}

\section{PROPOSED DEEP REINFORCEMENT LEARNING FRAMEWORK}

This section presents the DRL framework, and reward design for quick, efficient and collision-free AGV exploration in new, unknown environments. 

\subsection{Network Architecture}

\begin{figure}[!h]
    \centering
    \includegraphics[width = 1.0\columnwidth]{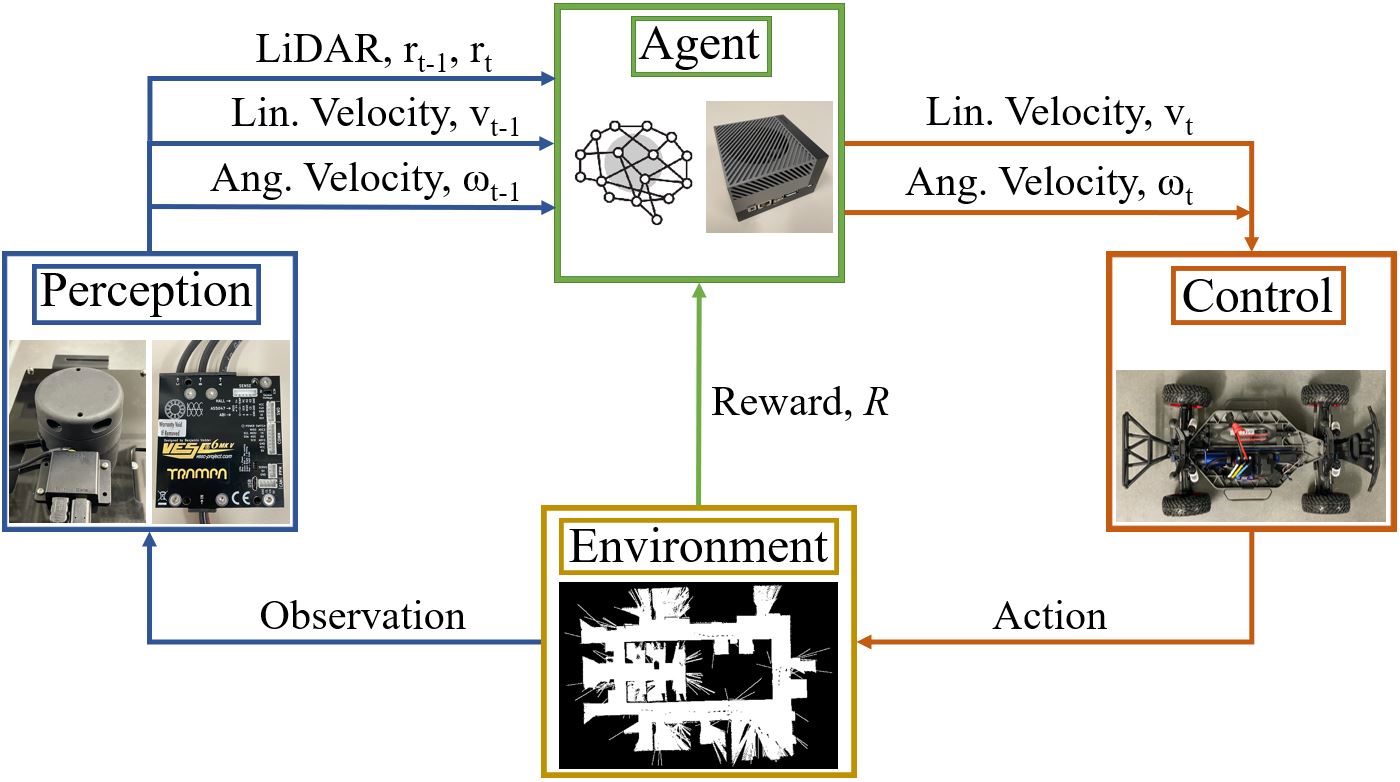}
    \caption{Deep Reinforcement Learning framework for Autonomous Ground Vehicle exploration without maps.} \label{arch}
\end{figure}

In order to maximize the long-term reward, designed to encourage quick, efficient, and collision-free exploration of the environment, the DRL agent makes strategic linear and angular velocity action decisions for the current time step, $v_{t}$ and $\omega_{t}$. These decisions are based on LiDAR or camera RGB-D point cloud range measurements from the current and previous time steps $r_{t}$ and $r_{t-1}$, the previous time step's action $a_{t-1} = (v_{t-1}, \omega_{t-1})$, and the corresponding reward value, $R$. The proposed DRL architecture for AGV exploration is shown in Figure \ref{arch}.

The agent updates the policy at each time step during training using the selected action, the prior and current observations after executing the action, and scalar reward feedback, with the objective of maximizing the long-term reward to promote collision-free exploration.

The addition of global position information obtained via GPS to the observation space will improve learning sample efficiency, and can encourage more efficient exploration, however it is not viable in environments that are underground, or inside buildings. To mitigate this, the reward is specifically shaped to encourage the agent to navigate previously unexplored regions solely utilizing range and odometry measurements.

\subsection{Reward Shaping}

The reward function is application specific, and designed to encourage the agent to explore its environment efficiently, quickly and safely without collisions. To compare the exploration capabilities of DDPG, TD3 and SAC, a minimalist reward function is shaped as presented in \cite{sivashangaran2021intelligent},

\begin{equation} \label{eqn1}
\textit{R} = 0.0075r^{2} + 1.5v^{2} - 0.6\omega^{2}
\end{equation}\

A positive reward is applied to the square of the minimum range measurement, $r$ to incentivize obstacle avoidance. This reward is highest when the agent is at a greater distance from obstacles, encouraging the generation of paths devoid of obstacles. The agent is additionally rewarded for swift navigation through positive reinforcement of linear velocity, $v$. To encourage efficient exploration, a negative reward is applied to angular velocity, $\omega$ to discourage repeated circular motion in the same vicinity. High coefficients for $r^2$ and $v^2$ lead to a compromise between obstacle avoidance ability and exploratory behavior, hence a balance was determined through trial-and-error experimentation to prioritize both exploration, and collision avoidance.

The presented DRL framework can be trained with more specialized application specific reward functions, such as for SAR operations in forests with thick foliage or urban regions covered in rubble in the aftermath of a natural disaster that require AGVs to specifically explore shaded regions underneath trees, boulders or buildings inaccessible to aerial surveillance using Unmanned Aerial Vehicles (UAVs). The reward for this application can be shaped as,

\begin{equation} \label{eqn2}
    R_{SAR} = 
\begin{dcases}
    0.0075r^{2} + 1.5v^{2} - 0.6\omega^{2} &\\ 
    2.0              & \text{if } 2.0 < r < 2.5\\
    -50              & \text{if } r < 1.0\\
\end{dcases}
\end{equation}

In addition to the reward in Equation \ref{eqn1}, the agent is assigned a reward of 2.0 when it is between 2.0 and 2.5 $m$ of the nearest object to promote exploration in shaded regions close to obstacles unobservable to aerial surveillance. In order to improve the safety of exploration trajectories, the agent is assigned a penalty of 50.0 when it is within 1 $m$ of the nearest obstacle.

The focus of this paper is on evaluating the proposed DRL framework for end-to-end cognitive AGV exploration without maps, and comparing the performance of sample efficient off-policy algorithms for this task, hence we use the generalized reward function in Equation \ref{eqn1} for comparison analysis. The best algorithm is further evaluated using the application specific shaped reward in Equation \ref {eqn2} to demonstrate the effectiveness of the proposed method.

\section{TRAINING AND EVALUATION}

The MATLAB Robotics System \cite{matlabRob} and Reinforcement Learning \cite{matlabRL} Toolboxes, and Simulink are utilized to model the AGV, and train the DRL agent to compare off-policy algorithms DDPG, TD3 and SAC using the reward in Equation \ref{eqn1}. The best algorithm is further evaluated in AutoVRL \cite{sivashangaran2023autovrl}, an in-house high fidelity simulator developed using open-source tools for simulation to real-world DRL using the shaped reward in Equation \ref{eqn2}.

\subsection{Environments}

\begin{figure}[htbp]
\centering
\begin{minipage}{0.25\textwidth}
\includegraphics[width=1.0\textwidth]{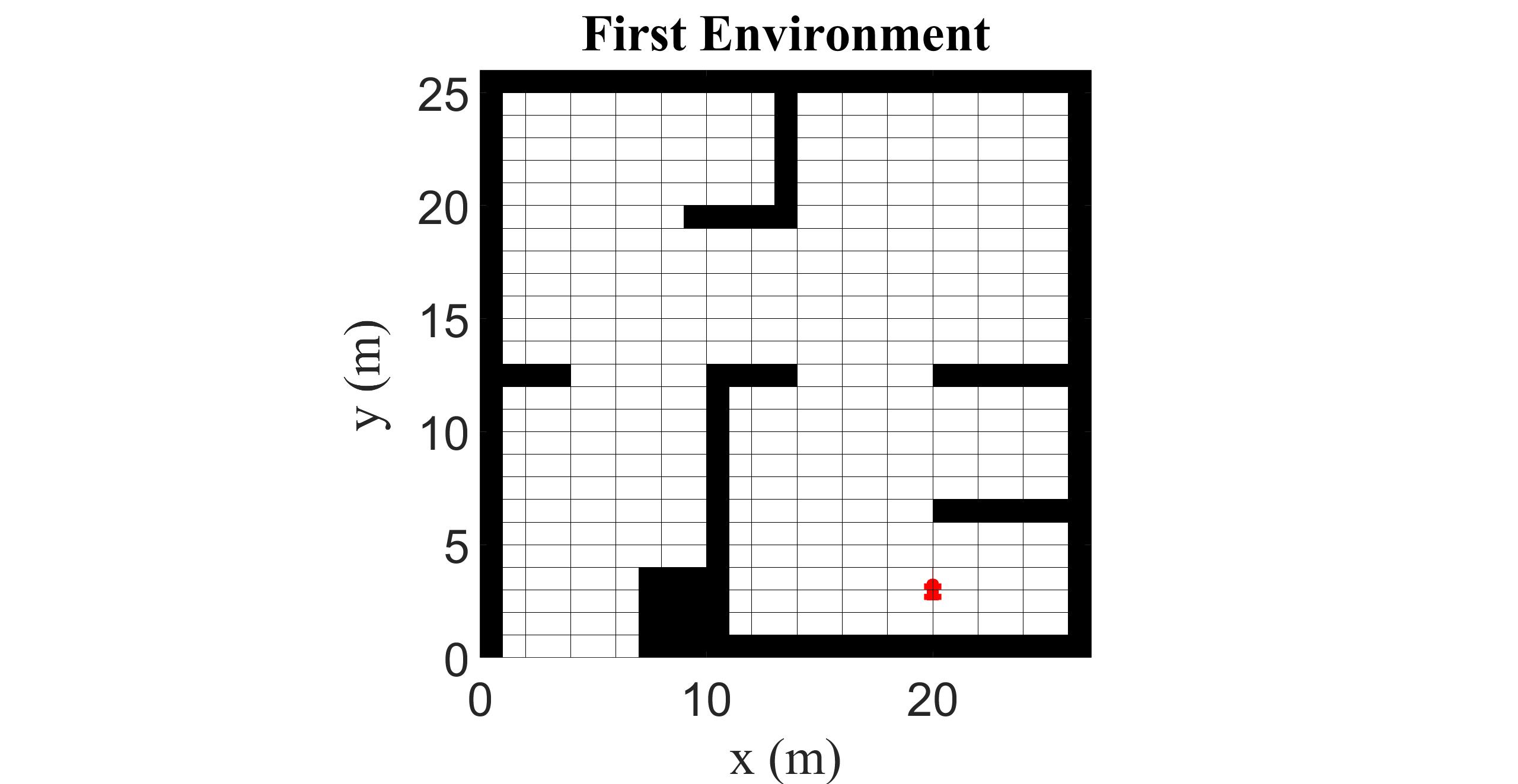}
\subcaption{} \label{trainenv1}
\end{minipage}%
\begin{minipage}{0.25\textwidth}
\includegraphics[width=1.0\textwidth]{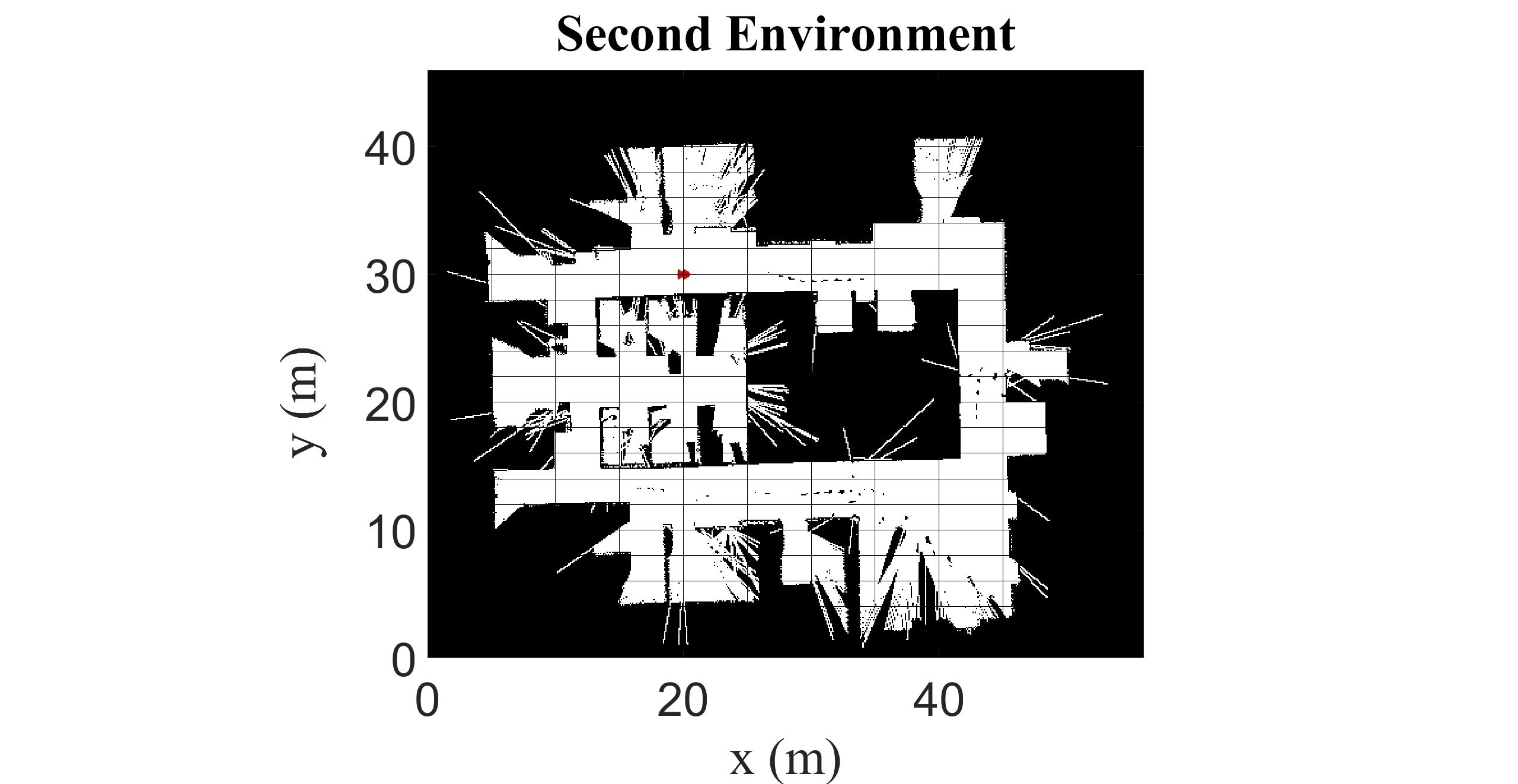}
\subcaption{} \label{trainenv2}
\end{minipage}%
\\
\begin{minipage}{0.25\textwidth}
\includegraphics[width=1.0\textwidth]{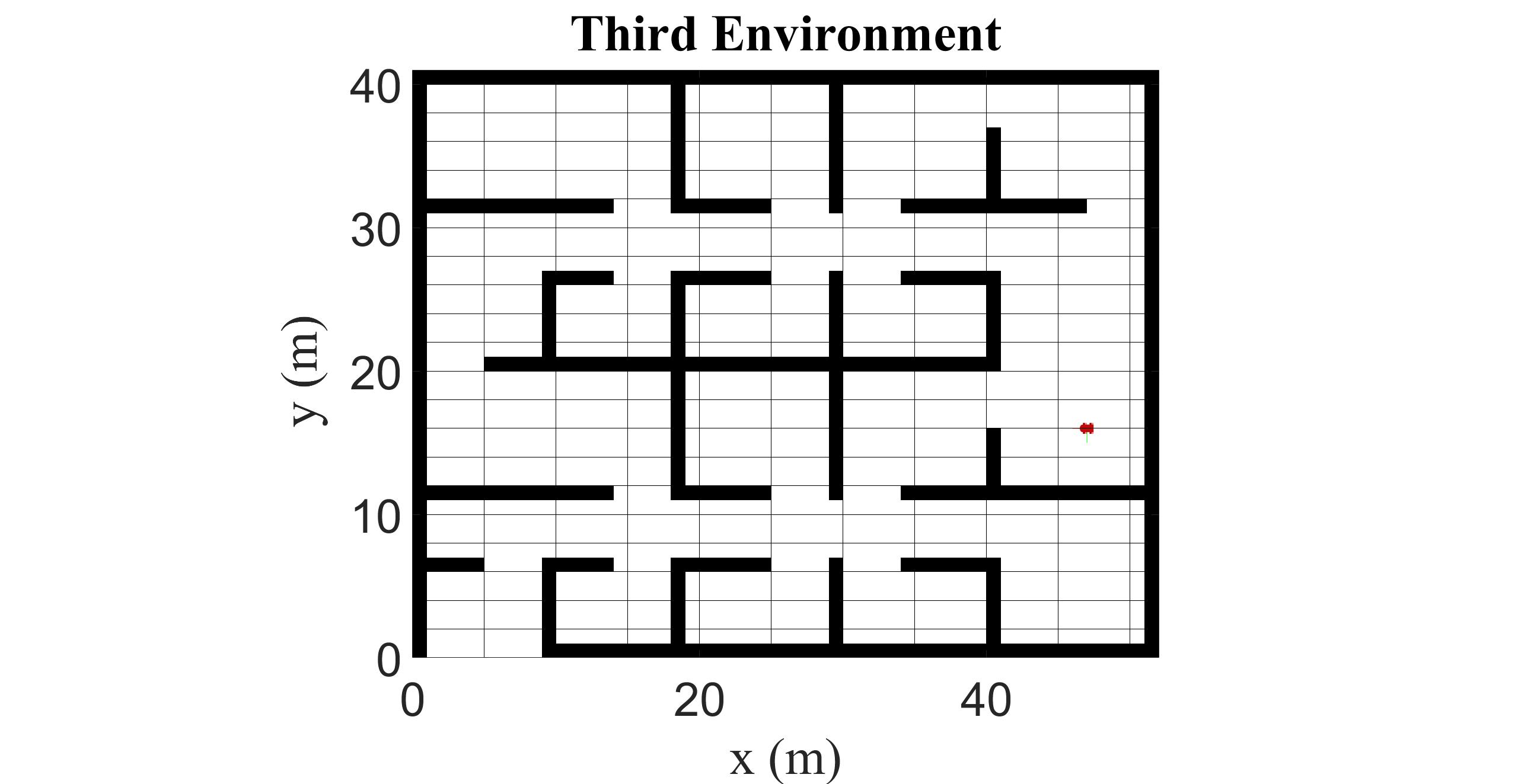}
\subcaption{} \label{evalenv}
\end{minipage}%
\caption{Training and evaluation environments with DRL agent marked in red at a randomized initial location. (a) First training environment. (b) Second training environment. (c) Evaluation environment.} \label{envs}
\end{figure}

The DRL agents are trained in two distinct environments for comparison. The first environment, depicted in Figure \ref{trainenv1}, is a simple 25 \textit{m} x 25 \textit{m} space with walls that the agent must steer clear of. The second environment, illustrated in Figure \ref{trainenv2}, is a more complex 40 \textit{m} x 40 \textit{m} space with walls and various obstacles, dotted in black, that the agent must additionally avoid. The trained agents from each environment are evaluated in a third environment, illustrated in Figure \ref{evalenv} without prior training or knowledge of map characteristics to evaluate the robustness, and performance of the learned policies in a new, unknown environment with the same network architecture and hyperparameters.

The AGV, identified with a red symbol on the training maps, is set to a random starting position at the start of each training episode to enhance policy learning. This reset ensures that the agent is not biased towards any particular initial location.

\begin{figure}[!h]
    \centering
    \includegraphics[width = 0.5\columnwidth]{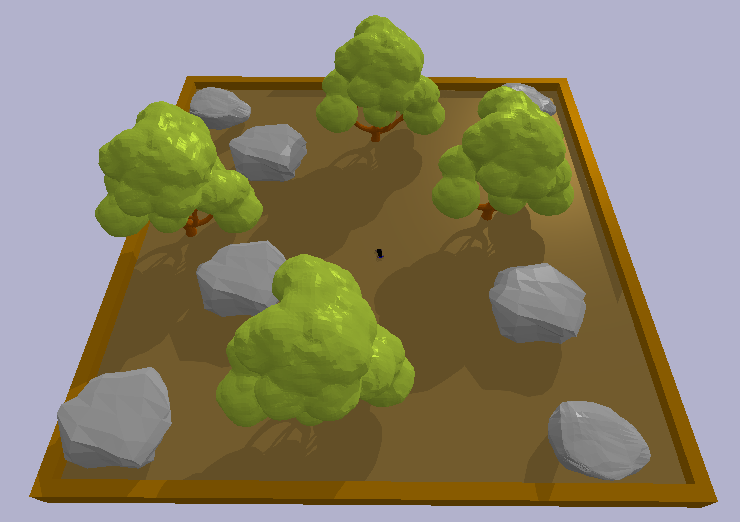}
    \caption{Outdoor environment with tree and boulder objects.} \label{autovrl_env20}
\end{figure}

To further evaluate the effectiveness of the DRL architecture for intelligent application specific exploration, a realistic 3D 20 \textit{m} x 20 \textit{m} outdoor environment with tree and boulder objects, illustrated in Figure \ref{autovrl_env20}, is utilized to train and evaluate the best algorithm determined from analyses of prior results.

This environment includes shaded regions beneath trees and boulders unobservable to aerial surveillance, hence the shaped reward in Equation \ref{eqn2} is utilized to train the DRL agent to specifically explore shaded regions inaccessible to UAVs.

\subsection{Exploration Quality and Efficiency Quantification}

\begin{figure}[!h]
    \centering
    \includegraphics[width = \columnwidth]{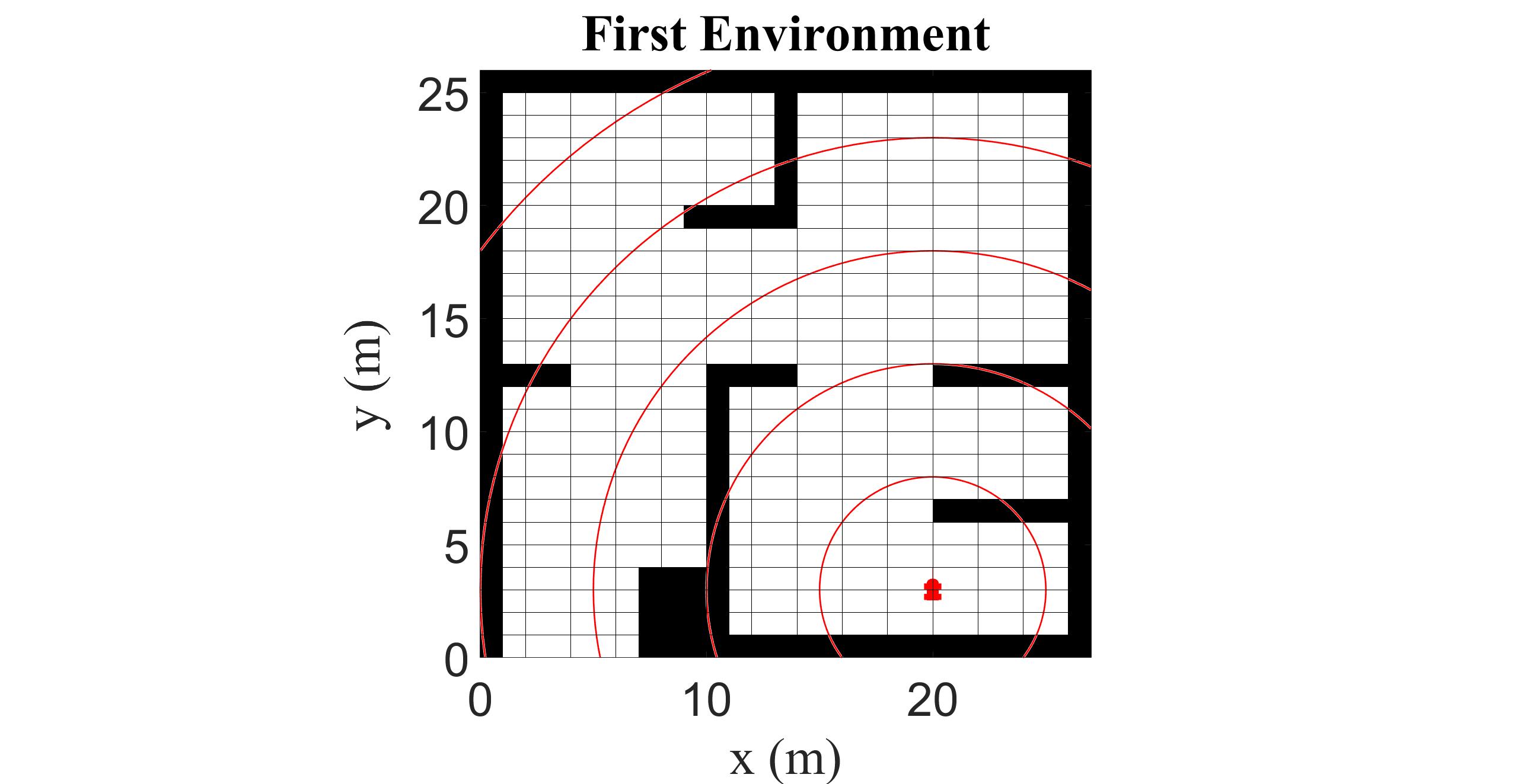}
    \caption{Occupancy grid map of first environment segmented in circles of increasing radii to score exploration quality and efficiency.} \label{env1_segment}
\end{figure}

The post-training trajectories learned by the agent are evaluated for exploration quality and efficiency by assigning an Exploration Quality Score (EQS) and Exploration Efficiency Score (EES) to quantitatively compare the performance of each tested algorithm. The training and evaluation environments are segmented as shown in Figure \ref{env1_segment} for the first training environment. Each environment is divided into 2 $m^{2}$ segments, and further segmented by circles of increasing radii with radius $10n$, where $n = 1, 2, 3, ...$ 

The EQS for the generated trajectory is scored as follows,

\begin{equation} \label{}
EQS = \Sigma_{n=1}^{n_{max}}(n\Sigma_{k=1}^{k_{max}}k)
\end{equation}

Here k is the number of 2 $m^{2}$ segments covered by the trajectory. The EQS is greater when the trajectory covers a higher number of segments, and traverses further away from the AGV's initial position to regions encompassed by circles of larger radii. 2 $m^{2}$ segments were chosen since it is reasonable for the AGV's onboard sensors to collect viable data for down-stream applications within this region.

The EES is scored as follows where $d$ is the trajectory distance,

\begin{equation} \label{}
EES = \frac{EQS}{d}
\end{equation}

The EES is higher when the EQS is large, and the distance travelled is small. A higher EES indicates better energy efficiency, and exploration performance.

\subsection{AGV Model}

XTENTH-CAR \cite{sivashangaran2022xtenth}, a proportionally scaled experimental vehicle platform, designed with similar hardware and software architectures as the full-size X-CAR \cite{mehr2022XCAR} connected autonomous vehicle, was modeled and trained in simulation. The XTENTH-CAR AGV has a wheelbase of 0.32 $m$ and utilizes the Ackermann steering mechanism.

\begin{figure}[thpb]
      \centering
      \includegraphics[width=0.6\columnwidth]{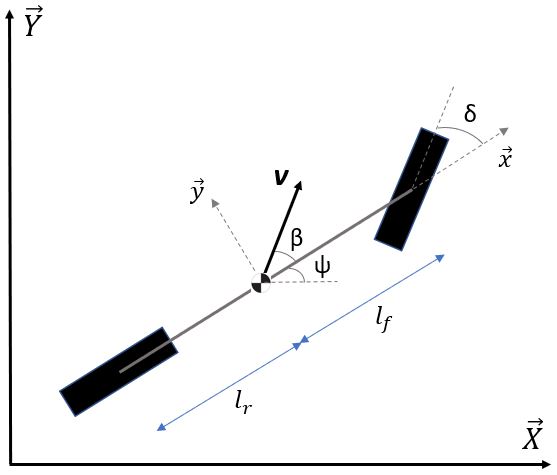}
      \caption{Schematic of kinematic bicycle model.}
      \label{bicycleModel}
\end{figure}

The AGV's kinematics are computed using a bicycle model, portrayed in Figure \ref{bicycleModel}, where the front and rear wheels are represented by a single wheel located at the center of each axle. This model is accurate for use at low speeds and offers a good balance between model accuracy and computation cost \cite{polack2017kinematic} for evaluation of the DRL agent.

The bicycle model is represented by the following equations,

\begin{equation} \label{}
\dot{x} = \textit{v}\,cos(\psi + \beta) 
\end{equation}\
\begin{equation} \label{}
\dot{y} = \textit{v}\,sin(\psi + \beta)
\end{equation}\
\begin{equation} \label{}
\dot{\psi} = \frac{\textit{v}}{l_{r}}\,sin(\beta) 
\end{equation}\
\begin{equation} \label{}
\beta = tan^{-1}\left(\frac{l_{r}}{l_{f}+l_{r}}\,tan(\delta)\right)
\end{equation}\

Here \textit{x} and \textit{y} are position coordinates of the AGV's center of mass, $\psi$ is the angle of the AGV's heading with respect to the inertial reference frame, $\beta$ is the angle between the velocity vector of the AGV's center of mass and its longitudinal axis, $l_{f}$ and $l_{r}$ are distances from the center of mass to the front and rear axles respectively, and velocity, \textit{v} and steering angle, $\delta$ are control inputs.

\subsection{Training Conditions}

A training episode is concluded when the agent encounters an obstacle or completes the maximum number of steps permitted in a single episode. Subsequently, the agent is reset to a randomly determined starting location to initiate the next episode.

The DRL agent is trained to a total of 10,000 episodes, each with a maximum of 1000 steps in the first environment, and 20,000 episodes, each with a maximum of 2000 steps in the second, to facilitate rapid iterative learning. The best algorithm is trained for 5000 episodes, each with a maximum of 1000 steps in the 3D outdoor environment. 

Hyperparameters that were modified with non-default values are listed in Table \ref{Hyperparameters}.

\begin{table}[ht]
    \renewcommand{\arraystretch}{1.2}
    \centering
    \caption{Non-Default Hyperparameters}
    \label{Hyperparameters}
        \begin{tabular}[t]{|c|c|}
            \hline
            \textbf{Hyperparameter} & \textbf{Value} \\
            \hline
            Discount Factor ($\gamma$)  & 0.995 \\
            Actor Learn Rate & 0.00005 \\
            Critic Learn Rate & 0.0005 \\
            Target Smooth Factor & 0.001 \\
            Mini Batch Size & 128 \\
            Experience Buffer Length & 1,000,000 \\
            \hline
        \end{tabular}
\end{table}

\section{RESULTS AND DISCUSSION}

This section presents DRL training results that include post-training exploration trajectories and corresponding average return and steps achieved by the agent each episode iteration during the training period, utilizing DDPG, TD3 and SAC algorithms. 

\subsection{Training Performance}

An Intel i7 11700K CPU and GeForce RTX 3070 Ti GPU were used for training. Table \ref{TrainTime} summarizes the training times for each DRL algorithm in the evaluated environments.

\begin{table}[ht]
    \renewcommand{\arraystretch}{1.2}
    \centering
    \caption{Training Times}
    \label{TrainTime}
        \begin{tabular}[t]{|c|c|}
            \hline
            \textbf{Agent} & \textbf{Training Time (Hrs)} \\
            \hline
            First Environment: DDPG  & 66.0 \\
            First Environment: TD3 & 95.3 \\
            First Environment: SAC & 156.7 \\
            Second Environment: DDPG & 84.8 \\
            Second Environment: TD3 & 128.1 \\
            Second Environment: SAC & 203.9 \\
            \hline
        \end{tabular}
\end{table}

SAC required the longest training time, followed by TD3 and DDPG which required the least. On average, training in the second, more complex environment required 31\% longer training time than in the first, over twice the number of training episodes. DDPG required 28.5\%, TD3 34.4\% and SAC 30.1\% longer to train in the second environment.

In the first environment, TD3 required 44.4\% longer to train than DDPG, and SAC 137.4\% longer than DDPG and 64.4\% longer than TD3. In the second environment, TD3 required 51.1\% longer to train than DDPG, and SAC 140.4\% longer than DDPG and 59.2\% longer than TD3. On average, TD3 required 47.8\% longer training time than DDPG, and SAC 138.9\% longer than DDPG and 61.8\% longer training time than TD3.

Training times ranged from 2.75 days in the first environment for DDPG to 8.5 days in the second environment for SAC. More optimal policies require longer training times to accommodate increased episode steps in the first environment, and more training episodes in the second.

\subsubsection{First Environment}

\begin{figure}[!h]
    \centering
    \includegraphics[width = \columnwidth]{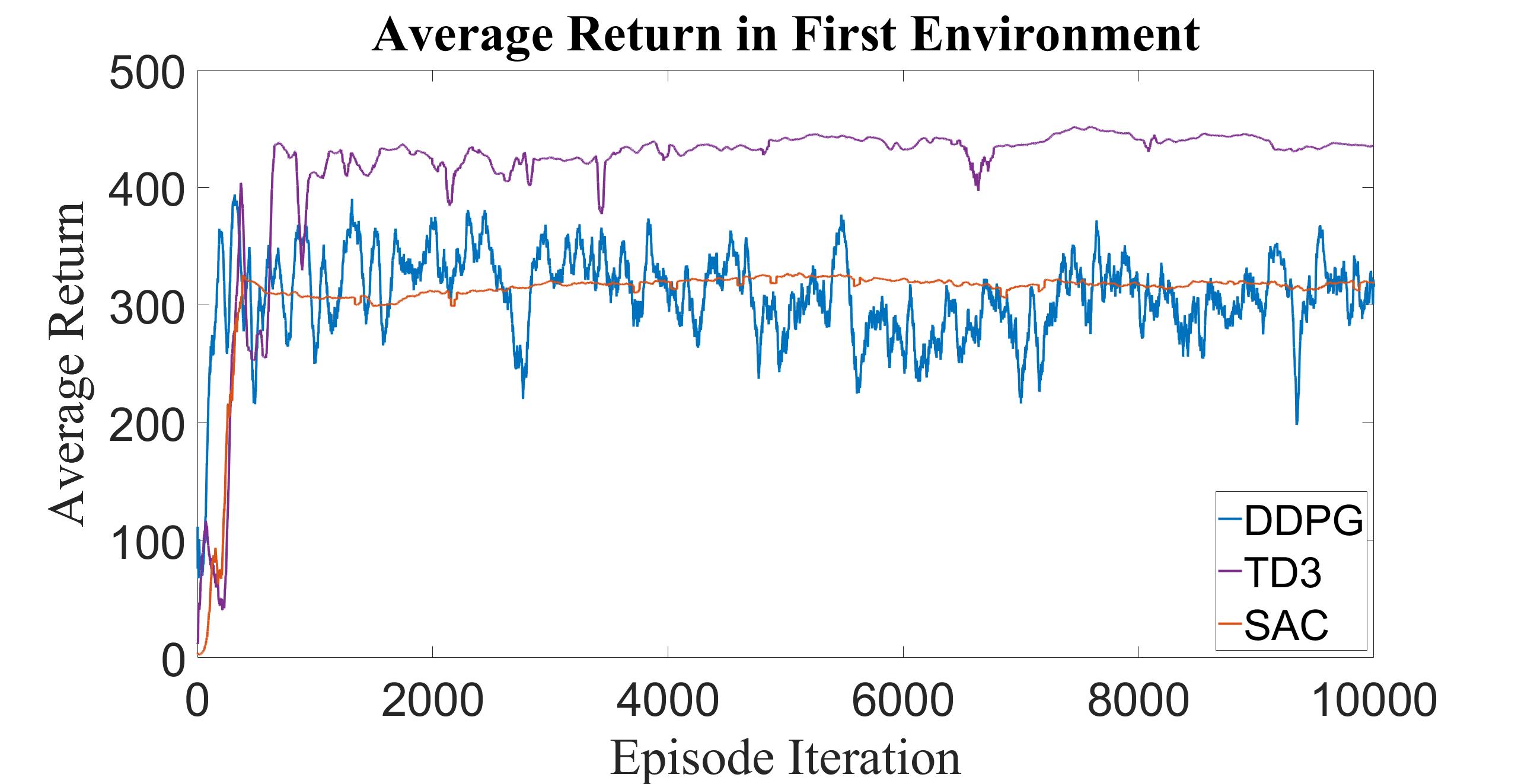}
    \caption{Order 50 moving average return during training in the first environment.} \label{avgReturnFirstEnv}
\end{figure}

\begin{figure}[!h]
    \centering
    \includegraphics[width = \columnwidth]{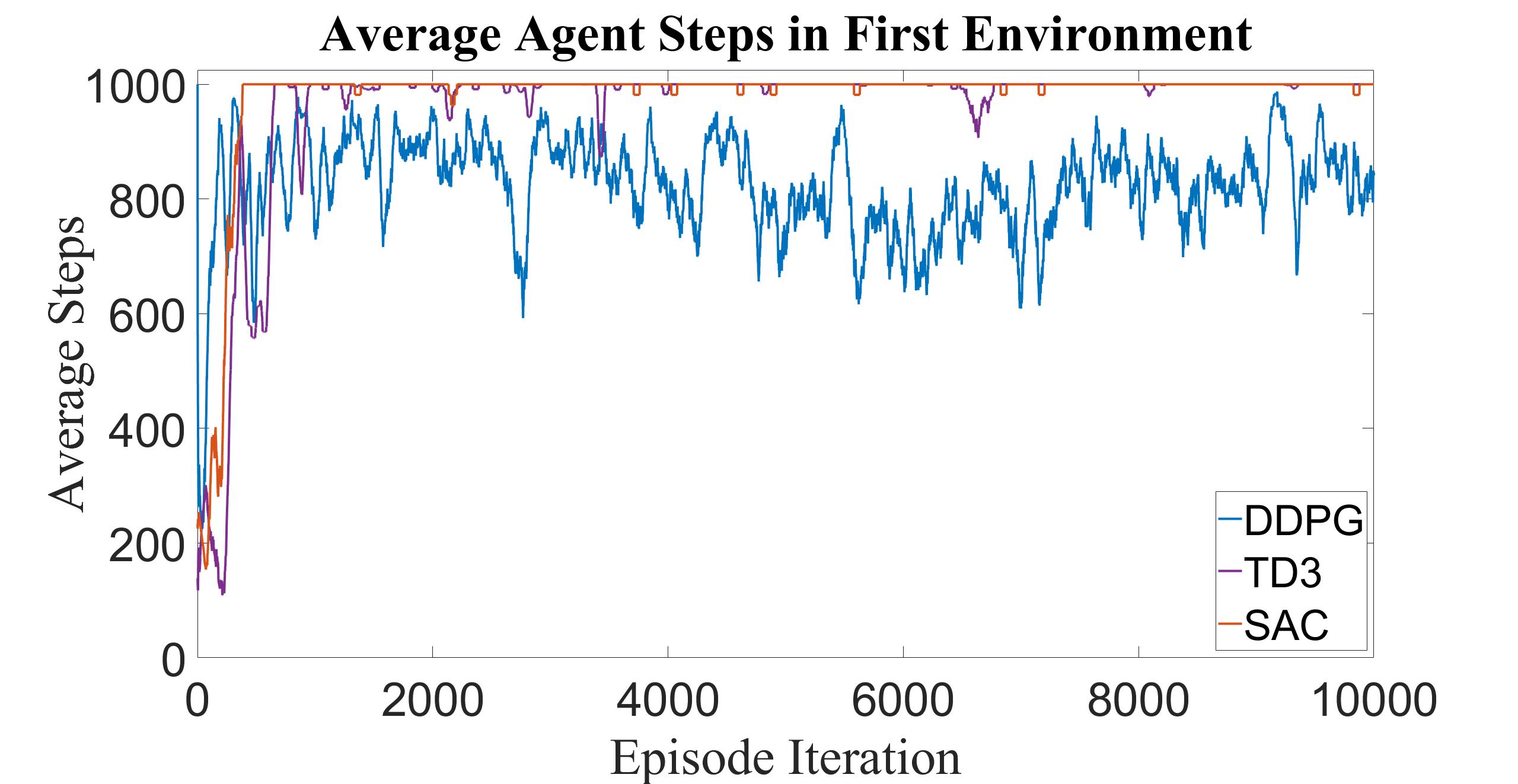}
    \caption{Order 50 moving average agent steps during training in the first environment.} \label{avgStepsFirstEnv}
\end{figure}

The order 50 moving average return and agent steps during training in the first environment are illustrated in Figures \ref{avgReturnFirstEnv} and \ref{avgStepsFirstEnv}. The training results in the first environment are summarized in Table \ref{firstEnvResults}.

\begin{table}[ht]
    \renewcommand{\arraystretch}{1.2}
    \centering
    \caption{First Environment Training Results}
    \label{firstEnvResults}
        \begin{tabular}[t]{|c|c|c|c|c|c|}
            \hline
            \textbf{Algorithm} & \textbf{Convergence} & \textbf{Average} & \textbf{Average} & \textbf{EQS} & \textbf{EES}\\
            & \textbf{Episode} & \textbf{Return} & \textbf{Steps} & & \\
            \hline
            DDPG & 170 & 318 & 865 & 81 & 1.82 \\
            TD3 & 960 & 435 & 1000 & \textbf{113} & 2.40 \\
            SAC & 390 & 320 & 1000 & 91 & \textbf{2.63}\\
            \hline
        \end{tabular}
\end{table}

DDPG converged first at 170 episodes with an average return of 318 and 865 average steps. TD3 converged last at 960 episodes with an average return of 435 and 1000 average steps, and SAC converged at 390 episodes with an average return of 320 and the maximum 1000 average steps. TD3 achieved the highest EQS, and SAC the highest EES. 

DDPG learned the least optimal policy with the lowest average return, agent steps, EQS and EES. TD3 achieves the highest return, and the maximum 1000 steps, however SAC achieves 1000 exploration steps more consistently post training convergence. Unlike TD3 which solely maximizes the long-term expected reward, SAC additionally maximizes the entropy of the policy to promote exploration. Consequently, TD3 learns a policy with a higher return, but SAC learns the better policy for agent exploration.



\begin{figure}[!h]
    \centering
    \includegraphics[width = \columnwidth]{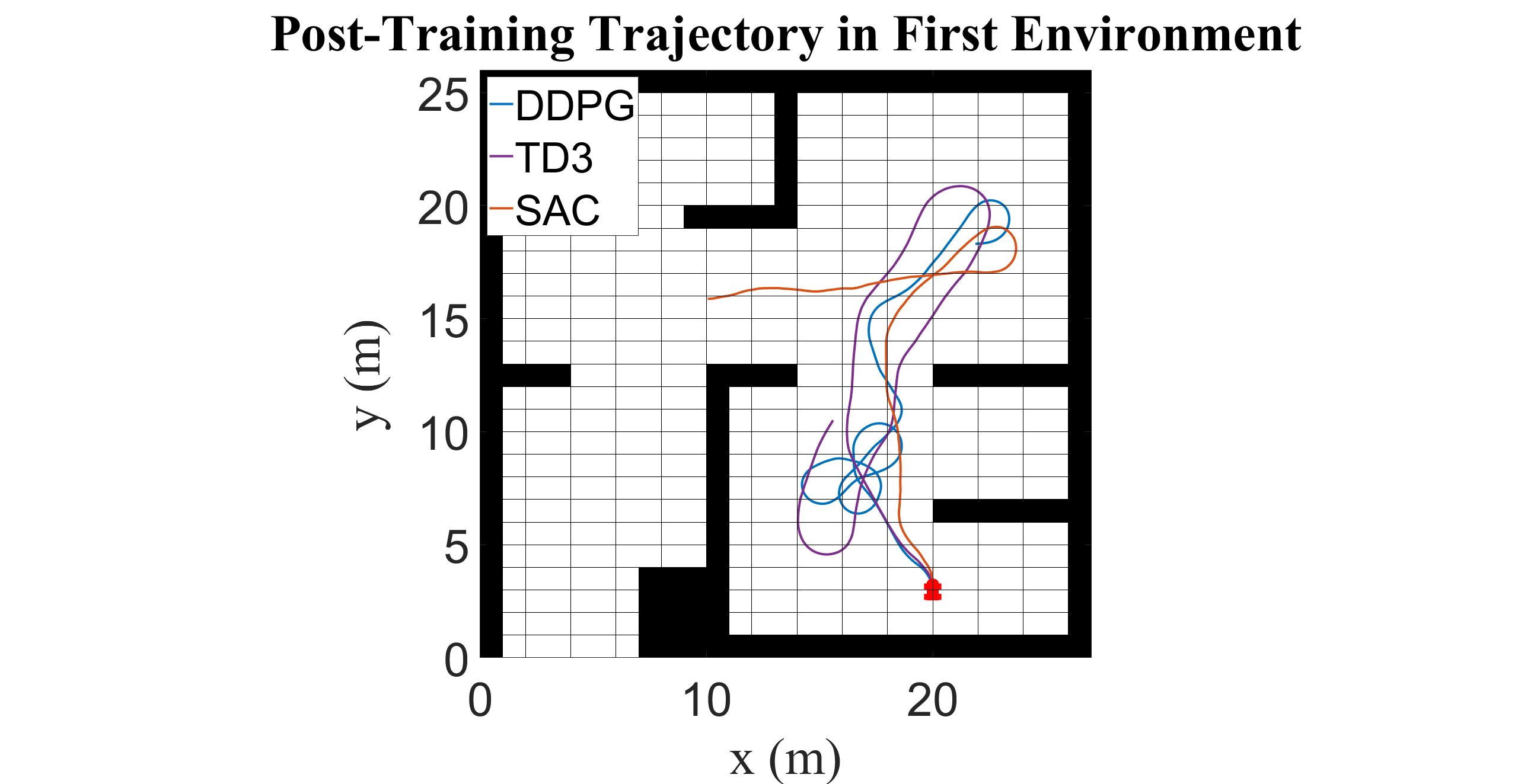}
    \caption{Trajectories in the first environment post training completion.} 
    \label{postTrajFirstEnv}
\end{figure}

 
The trajectories in the first environment for each algorithm post training completion are illustrated in Figure \ref{postTrajFirstEnv}.

Each algorithm achieved 1000 episode steps without collision. SAC covered the most ground, and exhibited the most efficient exploratory behavior which will result in the greatest energy savings. TD3 is next best, followed by DDPG which was the most inefficient, covering the same region multiple times. 

\subsubsection{Second Environment}

\begin{figure}[!h]
    \centering
    \includegraphics[width = \columnwidth]{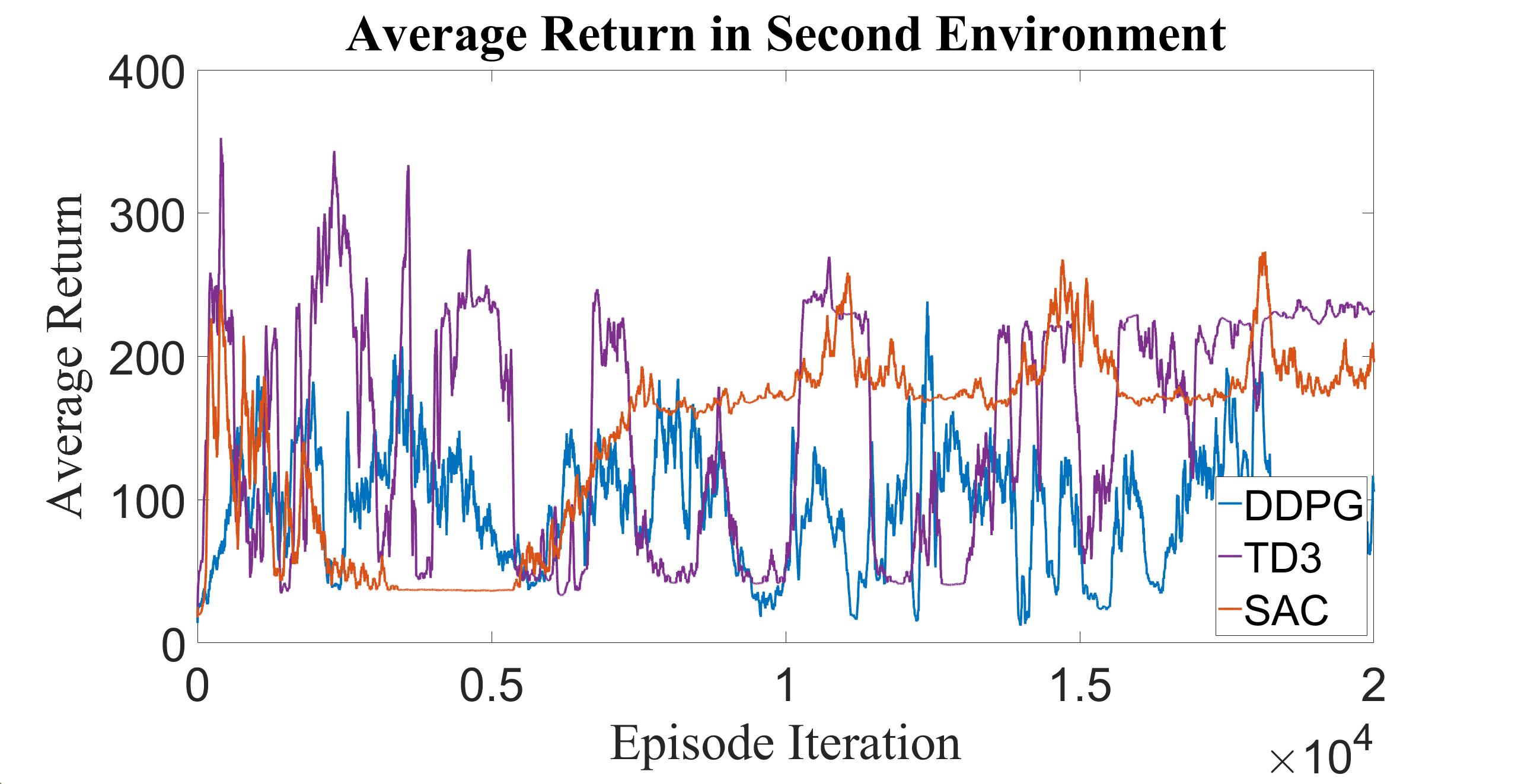}
    \caption{Order 50 moving average return during training in the second environment.} \label{avgReturnSecondEnv}
\end{figure}

\begin{figure}[!h]
    \centering
    \includegraphics[width = \columnwidth]{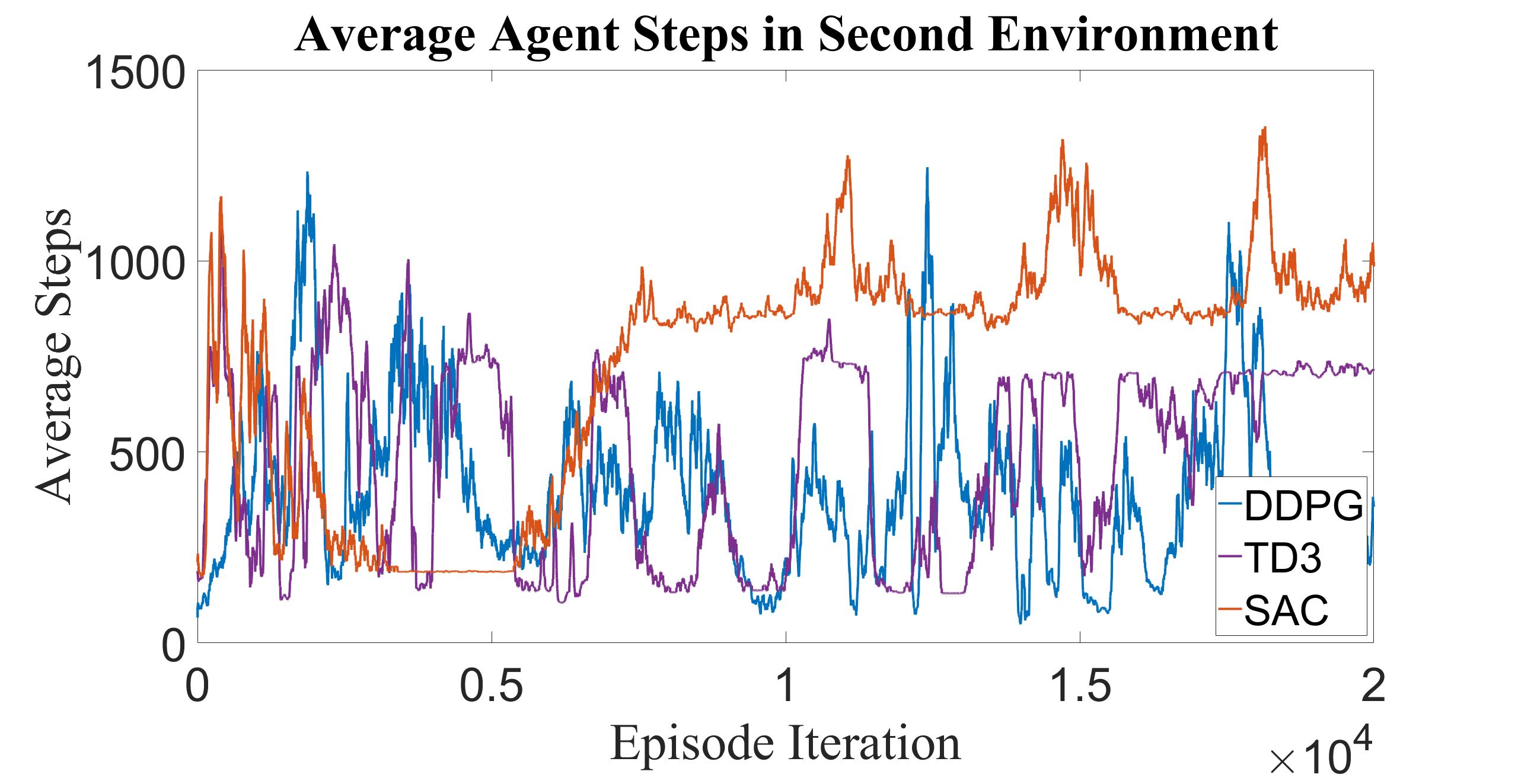}
    \caption{Order 50 moving average agent steps during training in the second environment.} \label{avgStepsSecondEnv}
\end{figure}

The order 50 moving average return and agent steps during training in the second environment are illustrated in Figures \ref{avgReturnSecondEnv} and \ref{avgStepsSecondEnv}. The training results in the second environment are summarized in Table \ref{secondEnvResults}.

Training for 20,000 episodes was insufficient for the DRL algorithms to learn an optimal policy in the second environment. At the end of the training period, DDPG achieved an average return of 125 and 530 average steps, TD3 obtained an average return of 230 and 715 average steps, and SAC converged to a local maximum at 10,620 episodes with an average return of 210 and 1050 average steps. Training was limited to 20,000 episodes to gauge performance in a reasonable time frame, however, continued training over 75,000 to 100,000 episodes will enable the agents to learn an optimal policy to traverse the more complex terrain over an indefinite number of exploration steps. 

\begin{table}[ht]
    \renewcommand{\arraystretch}{1.2}
    \centering
    \caption{Second Environment Training Results}
    \label{secondEnvResults}
        \begin{tabular}[t]{|c|c|c|c|c|}
            \hline
            \textbf{Algorithm} & \textbf{Average} & \textbf{Average} & \textbf{EQS} & \textbf{EES}\\
            &  \textbf{Return} & \textbf{Steps} & & \\
            \hline
            DDPG & 125 & 530 & 60 & 1.92\\
            TD3  & 230 & 715 & 67 & 1.97\\
            SAC  & 210 & 1050 & \textbf{110} & \textbf{2.24}\\
            \hline
        \end{tabular}
\end{table}

Training DDPG, TD3 and SAC algorithms in the second environment for 20,000 episodes required a total of 416.8 hours, as such, it is infeasible to evaluate the algorithms for 75,000+ episodes with the existing setup. More powerful computer hardware is required. Similar to the training results in the first environment, DDPG learned the least optimal policy achieving the lowest return, agent steps, EQS and EES. TD3 achieved the highest return, however SAC learned a more optimal policy achieving the highest EQS and EES.  



\begin{figure}[!h]
    \centering
    \includegraphics[width = \columnwidth]{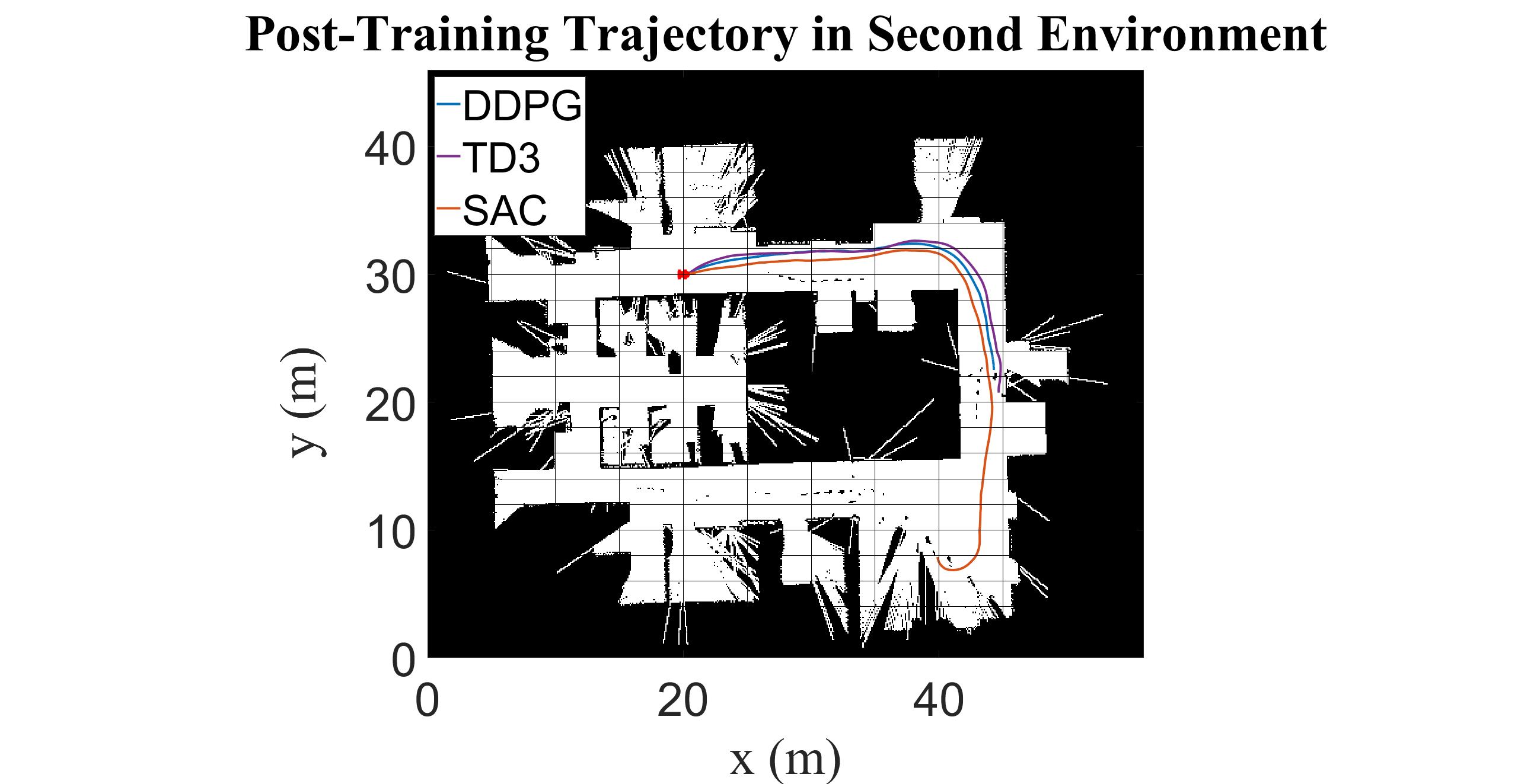}
    \caption{Trajectories in the second environment post training completion.} 
    \label{postTrajSecondEnv}
\end{figure}

 
The trajectories in the second environment for each algorithm post training completion are illustrated in Figure \ref{postTrajSecondEnv}.

SAC achieved the best performance, learning a trajectory that covers the most distance. TD3 and DDPG yield similar performance, with TD3 being a marginal improvement. 

\subsection{Trained Policy Evaluation}

\begin{figure}[!h]
    \centering
    \includegraphics[width = \columnwidth]{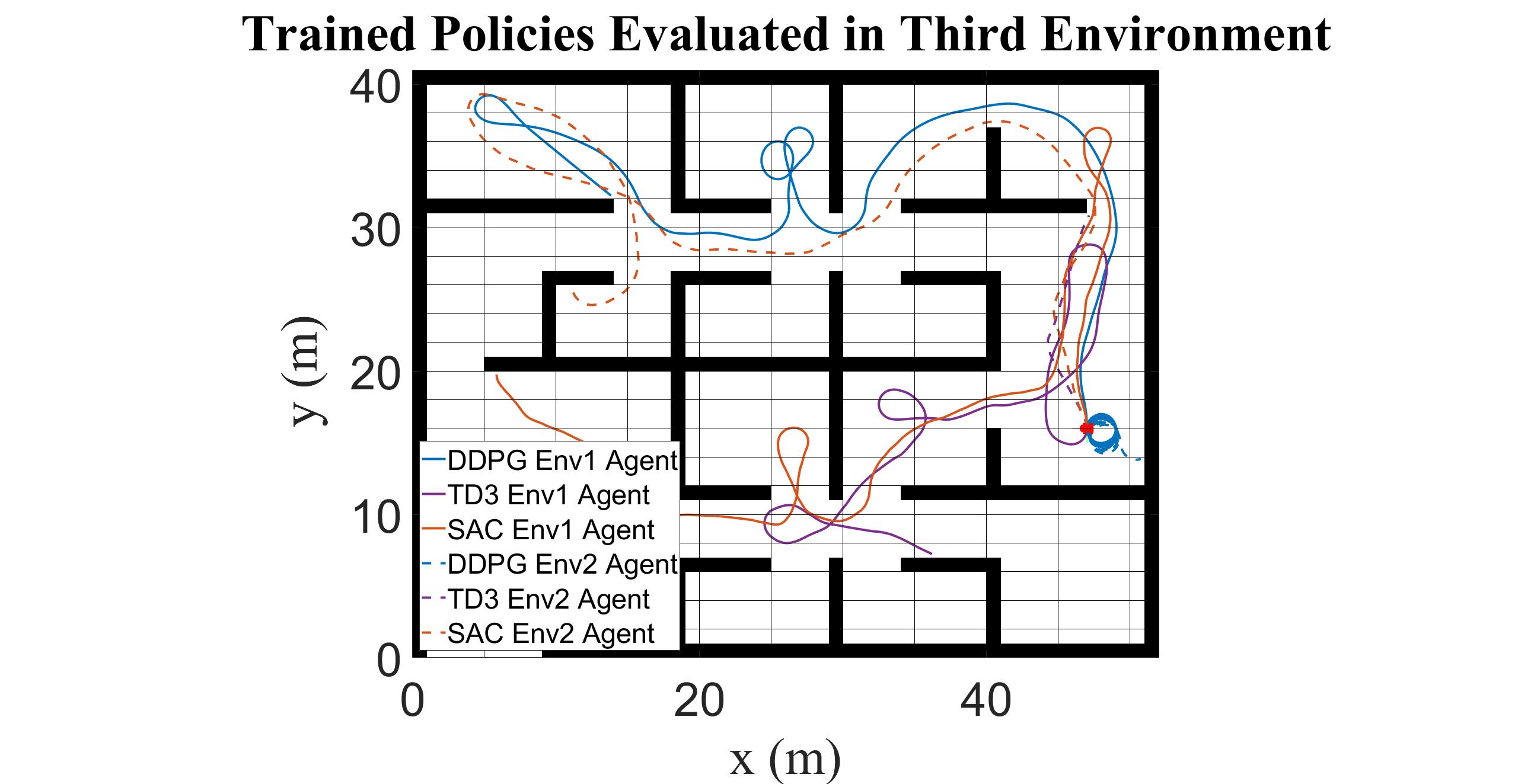}
    \caption{Trained DRL agents evaluated in the third environment.} 
    \label{thirdEnvEval}
\end{figure}

The six agents, DDPG, TD3 and SAC, trained in two different environments were evaluated in a third unknown environment with no prior training or knowledge of environment characteristics, to evaluate the extensibility of the ubiquitous DRL architecture for AGV exploration in information poor environments. Figure \ref{thirdEnvEval} portrays the trajectories for each agent in the third environment. The evaluation results are summarized in Table \ref{thirdEnvResults}.

\begin{table}[ht]
    \renewcommand{\arraystretch}{1.2}
    \centering
    \caption{Third Environment Evaluation Results}
    \label{thirdEnvResults}
        \begin{tabular}[t]{|c|c|c|c|}
            \hline
            \textbf{Agent} & \textbf{Steps} & \textbf{EQS} & \textbf{EES}\\
            \hline
            DDPG Env1 & 2151 & 252 & 2.36\\
            TD3 Env1 & 1533 & 102 & 1.33\\
            SAC Env1 & 2971 & 197 & 1.95\\
            DDPG Env2 & 2893 & 3 & 0.02\\
            TD3 Env2 & 321 & 25 & 1.56\\
            SAC Env2 & 2839 & \textbf{303} & \textbf{3.15}\\
            \hline
        \end{tabular}
\end{table}

The SAC agent trained in the second, more complex environment demonstrated the best performance, covering the most ground, efficiently with the highest EQS and EES. The DDPG agent trained in the first environment performed second best, covering more ground than either TD3 agent. DDPG trained in the second environment covered the second highest distance, but yielded the worst exploratory behavior with the least EQS and EES, repeatedly traversing a circular trajectory in the same vicinity. TD3 agents covered less ground, and exhibited less efficient exploratory behavior than either SAC agent. DDPG trained in the first environment performed better than that trained in the second, as the characteristics of the evaluated environment are more similar to the first than the second. The SAC agents are most robust to differences in environment characteristics, and performed better when trained in the more complex environment. Both SAC agents performed well, with the agent trained in the simple first environment achieving the third highest EQS and EES.

The reward function weights and network hyperparameters can be further engineered for this application, and the agent trained over a longer period with more episode steps each episode iteration to learn an improved policy that explores the surrounding environment indefinitely.

\subsection{Shaped Reward Evaluation}

\begin{figure}[!h]
    \centering
    \includegraphics[width = \columnwidth]{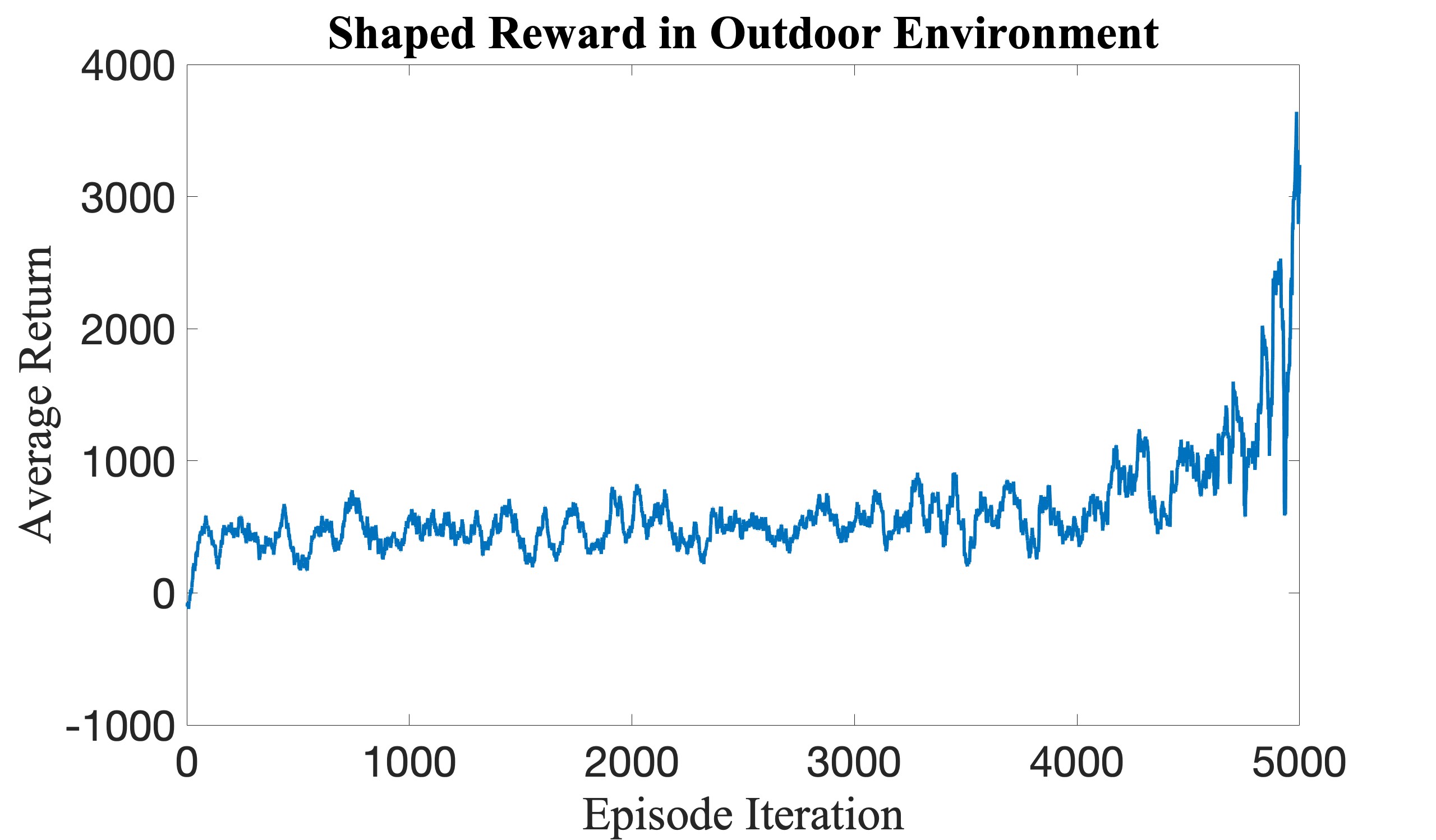}
    \caption{Order 50 moving average return during training in the outdoor environment.} 
    \label{shapedRewardPlot}
\end{figure}

\begin{figure}[!h]
    \centering
    \includegraphics[width = 0.8\columnwidth]{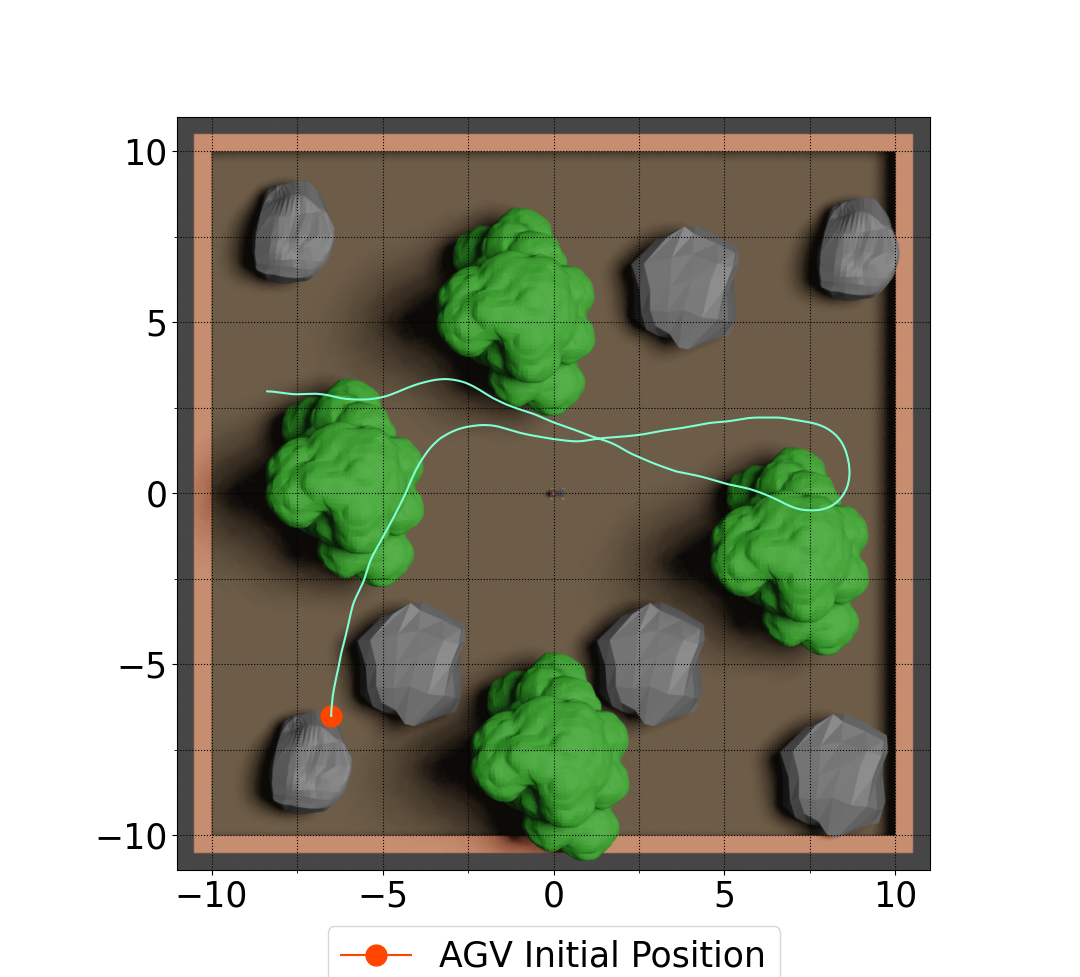}
    \caption{Trajectory in the outdoor environment post training completion.} 
    \label{3Denv}
\end{figure}

SAC was utilized to learn a policy in a realistic 3D environment using the shaped reward in Equation \ref{eqn2} designed for AGV exploration in shaded regions unobservable to UAVs. The order 50 moving average return during training is illustrated in Figure \ref{shapedRewardPlot}. A post-training trajectory from a randomized initial position is illustrated in Figure \ref{3Denv}.

The average return increased to 3642 at the end of the training period, yielding a policy that successfully promoted exploration in shaded regions beneath trees. 

Bridging the simulation to real-world gap to transfer policies learned in simulation to real-world robotic systems is a current area of active research. The large number of episodes required to sufficiently train the agent renders simulation training an essential component for DRL in robotics applications to minimize cost and possible physical damage caused by collisions during training. Substantial computation cost is required for training, however, post-training implementation of DRL agents is significantly less expensive, which makes DRL a powerful tool for real-time AGV motion planning and control in environments without a-priori maps.

In future work, AutoVRL will be utilized to learn policies to transfer to a physical XTENTH-CAR.

\section{CONCLUSIONS}

This paper presented an ubiquitous DRL architecture for intelligent AGV exploration without a-priori maps. Three actor-critic DRL algorithms, DDPG, TD3 and SAC, were trained in two environments of varying complexity, and further evaluated in a third with no prior knowledge of map characteristics. Simulation results demonstrate the effectiveness of the proposed DRL architecture, reward function and training conditions for quick, efficient and collision-free AGV navigation. SAC achieved the best performance, yielding trajectories that cover the highest distance, and demonstrated the most efficient exploratory behavior. Learning requires substantial computation cost, requiring up to 8.5 days for SAC in the second, complex environment using an Intel i7 11700K CPU and GeForce RTX 3070 Ti GPU. Improved policies with higher post-training episode steps require greater training times. Despite the high training cost, post-training implementation of DRL agents is significantly less expensive, which makes DRL a powerful tool for real-time AGV exploration in information poor, dynamically altering environments. Reward shaping renders the proposed DRL framework a versatile tool for a multitude of AGV applications. SAC was trained using a reward shaped for AGV exploration in regions inaccessible to aerial surveillance in a realistic outdoor environment, yielding a trajectory that specifically explored shaded regions underneath trees, demonstrating the effectiveness of the proposed DRL framework. In future work, the simulation to real-world gap will be bridged to transfer policies learned in simulation to the physical AGV using AutoVRL, a high fidelity DRL simulator developed using open-source tools.

\addtolength{\textheight}{-12cm}   






\bibliography{root}

\begin{thebibliography}{10}
\providecommand{\url}[1]{#1}
\csname url@samestyle\endcsname
\providecommand{\newblock}{\relax}
\providecommand{\bibinfo}[2]{#2}
\providecommand{\BIBentrySTDinterwordspacing}{\spaceskip=0pt\relax}
\providecommand{\BIBentryALTinterwordstretchfactor}{4}
\providecommand{\BIBentryALTinterwordspacing}{\spaceskip=\fontdimen2\font plus
\BIBentryALTinterwordstretchfactor\fontdimen3\font minus
  \fontdimen4\font\relax}
\providecommand{\BIBforeignlanguage}[2]{{%
\expandafter\ifx\csname l@#1\endcsname\relax
\typeout{** WARNING: IEEEtran.bst: No hyphenation pattern has been}%
\typeout{** loaded for the language `#1'. Using the pattern for}%
\typeout{** the default language instead.}%
\else
\language=\csname l@#1\endcsname
\fi
#2}}
\providecommand{\BIBdecl}{\relax}
\BIBdecl

\bibitem{sun2021motion}
H.~Sun, W.~Zhang, R.~Yu, and Y.~Zhang, ``Motion planning for mobile
  robots—focusing on deep reinforcement learning: A systematic review,''
  \emph{IEEE Access}, vol.~9, pp. 69\,061--69\,081, 2021.

\bibitem{victerpaul2017path}
P.~Victerpaul, D.~Saravanan, S.~Janakiraman, and J.~Pradeep, ``Path planning of
  autonomous mobile robots: A survey and comparison,'' \emph{Journal of
  Advanced Research in Dynamical and Control Systems}, vol.~9, no.~12, pp.
  1535--1565, 2017.

\bibitem{levinson2011towards}
J.~Levinson, J.~Askeland, J.~Becker, J.~Dolson, D.~Held, S.~Kammel, J.~Z.
  Kolter, D.~Langer, O.~Pink, V.~Pratt \emph{et~al.}, ``Towards fully
  autonomous driving: Systems and algorithms,'' in \emph{2011 IEEE intelligent
  vehicles symposium (IV)}.\hskip 1em plus 0.5em minus 0.4em\relax IEEE, 2011,
  pp. 163--168.

\bibitem{wang2011application}
H.~Wang, Y.~Yu, and Q.~Yuan, ``Application of dijkstra algorithm in robot
  path-planning,'' in \emph{2011 second international conference on mechanic
  automation and control engineering}.\hskip 1em plus 0.5em minus 0.4em\relax
  IEEE, 2011, pp. 1067--1069.

\bibitem{gammell2014informed}
J.~D. Gammell, S.~S. Srinivasa, and T.~D. Barfoot, ``Informed rrt*: Optimal
  sampling-based path planning focused via direct sampling of an admissible
  ellipsoidal heuristic,'' in \emph{2014 IEEE/RSJ International Conference on
  Intelligent Robots and Systems}.\hskip 1em plus 0.5em minus 0.4em\relax IEEE,
  2014, pp. 2997--3004.

\bibitem{vadakkepat2000evolutionary}
P.~Vadakkepat, K.~C. Tan, and W.~Ming-Liang, ``Evolutionary artificial
  potential fields and their application in real time robot path planning,'' in
  \emph{Proceedings of the 2000 congress on evolutionary computation. CEC00
  (Cat. No. 00TH8512)}, vol.~1.\hskip 1em plus 0.5em minus 0.4em\relax IEEE,
  2000, pp. 256--263.

\bibitem{bakdi2017optimal}
A.~Bakdi, A.~Hentout, H.~Boutami, A.~Maoudj, O.~Hachour, and B.~Bouzouia,
  ``Optimal path planning and execution for mobile robots using genetic
  algorithm and adaptive fuzzy-logic control,'' \emph{Robotics and Autonomous
  Systems}, vol.~89, pp. 95--109, 2017.

\bibitem{sivashangaran2021Thesis}
S.~Sivashangaran, ``Application of deep reinforcement learning for intelligent
  autonomous navigation of car-like mobile robot,'' Master's thesis, State
  University of New York at Buffalo, 2021.

\bibitem{sivashangaran2021intelligent}
S.~Sivashangaran and M.~Zheng, ``Intelligent autonomous navigation of car-like
  unmanned ground vehicle via deep reinforcement learning,''
  \emph{IFAC-PapersOnLine}, vol.~54, no.~20, pp. 218--225, 2021.

\bibitem{zhu2021deep}
K.~Zhu and T.~Zhang, ``Deep reinforcement learning based mobile robot
  navigation: A review,'' \emph{Tsinghua Science and Technology}, vol.~26,
  no.~5, pp. 674--691, 2021.

\bibitem{larsen2021comparing}
T.~N. Larsen, H.~{\O}. Teigen, T.~Laache, D.~Varagnolo, and A.~Rasheed,
  ``Comparing deep reinforcement learning algorithms’ ability to safely
  navigate challenging waters,'' \emph{Frontiers in Robotics and AI}, vol.~8,
  2021.

\bibitem{wijmansdd}
E.~Wijmans, A.~Kadian, A.~Morcos, S.~Lee, I.~Essa, D.~Parikh, M.~Savva, and
  D.~Batra, ``Dd-ppo: Learning near-perfect pointgoal navigators from 2.5
  billion frames,'' in \emph{8th International Conference on Learning
  Representations, ICLR 2020}, 2020.

\bibitem{chen2019learning}
T.~Chen, S.~Gupta, and A.~Gupta, ``Learning exploration policies for
  navigation,'' in \emph{7th International Conference on Learning
  Representations, ICLR 2019}, 2019.

\bibitem{cadena2016SLAMsurvey}
C.~Cadena, L.~Carlone, H.~Carrillo, Y.~Latif, D.~Scaramuzza, J.~Neira, I.~Reid,
  and J.~J. Leonard, ``Past, present, and future of simultaneous localization
  and mapping: Toward the robust-perception age,'' \emph{IEEE Transactions on
  Robotics}, vol.~32, no.~6, pp. 1309--1332, 2016.

\bibitem{ebadi2022present}
K.~Ebadi, L.~Bernreiter, H.~Biggie, G.~Catt, Y.~Chang, A.~Chatterjee, C.~E.
  Denniston, S.-P. Desch{\^e}nes, K.~Harlow, S.~Khattak \emph{et~al.},
  ``Present and future of {SLAM} in extreme underground environments,''
  \emph{arXiv preprint arXiv:2208.01787}, 2022.

\bibitem{chaplotlearning}
D.~S. Chaplot, D.~Gandhi, S.~Gupta, A.~Gupta, and R.~Salakhutdinov, ``Learning
  to explore using active neural slam,'' in \emph{8th International Conference
  on Learning Representations, ICLR 2020}.

\bibitem{schulman2017proximal}
J.~Schulman, F.~Wolski, P.~Dhariwal, A.~Radford, and O.~Klimov, ``Proximal
  policy optimization algorithms,'' \emph{arXiv preprint arXiv:1707.06347},
  2017.

\bibitem{lillicrap2015continuous}
T.~P. Lillicrap, J.~J. Hunt, A.~Pritzel, N.~Heess, T.~Erez, Y.~Tassa,
  D.~Silver, and D.~Wierstra, ``Continuous control with deep reinforcement
  learning,'' \emph{arXiv preprint arXiv:1509.02971}, 2015.

\bibitem{fujimoto2018addressing}
S.~Fujimoto, H.~Hoof, and D.~Meger, ``Addressing function approximation error
  in actor-critic methods,'' in \emph{International conference on machine
  learning}.\hskip 1em plus 0.5em minus 0.4em\relax PMLR, 2018, pp. 1587--1596.

\bibitem{haarnoja2018soft}
T.~Haarnoja, A.~Zhou, K.~Hartikainen, G.~Tucker, S.~Ha, J.~Tan, V.~Kumar,
  H.~Zhu, A.~Gupta, P.~Abbeel \emph{et~al.}, ``Soft actor-critic algorithms and
  applications,'' \emph{arXiv preprint arXiv:1812.05905}, 2018.

\bibitem{jones2007bat}
G.~Jones and M.~W. Holderied, ``Bat echolocation calls: adaptation and
  convergent evolution,'' \emph{Proceedings of the Royal Society B: Biological
  Sciences}, vol. 274, no. 1612, pp. 905--912, 2007.

\bibitem{sutton2018reinforcement}
R.~S. Sutton and A.~G. Barto, \emph{Reinforcement learning: An
  introduction}.\hskip 1em plus 0.5em minus 0.4em\relax MIT press, 2018.

\bibitem{silver2014deterministic}
D.~Silver, G.~Lever, N.~Heess, T.~Degris, D.~Wierstra, and M.~Riedmiller,
  ``Deterministic policy gradient algorithms,'' in \emph{International
  conference on machine learning}.\hskip 1em plus 0.5em minus 0.4em\relax PMLR,
  2014, pp. 387--395.

\bibitem{matlabRob}
\BIBentryALTinterwordspacing
{T}he~{M}athworks{,} {I}nc. {R}obotics {S}ystem {T}oolbox. MATLAB. Natick,
  Massachusetts, United States. [Online]. Available:
  \url{https://www.mathworks.com/products/robotics.html}
\BIBentrySTDinterwordspacing

\bibitem{matlabRL}
\BIBentryALTinterwordspacing
{The Mathworks, Inc.} {R}einforcement {L}earning {T}oolbox. MATLAB. Natick,
  Massachusetts, United States. [Online]. Available:
  \url{https://www.mathworks.com/products/reinforcement-learning.html}
\BIBentrySTDinterwordspacing

\bibitem{sivashangaran2023autovrl}
S.~Sivashangaran, A.~Khairnar, and A.~Eskandarian, ``Autovrl: A high fidelity
  autonomous ground vehicle simulator for sim-to-real deep reinforcement
  learning,'' \emph{arXiv preprint arXiv:2304.11496}, 2023.

\bibitem{sivashangaran2022xtenth}
S.~Sivashangaran and A.~Eskandarian, ``{XTENTH-CAR}: A proportionally scaled
  experimental vehicle platform for connected autonomy and all-terrain
  research,'' \emph{arXiv preprint arXiv:2212.01691}, 2022.

\bibitem{mehr2022XCAR}
G.~Mehr, P.~Ghorai, C.~Zhang, A.~Nayak, D.~Patel, S.~Sivashangaran, and
  A.~Eskandarian, ``{X-CAR}: An experimental vehicle platform for connected
  autonomy research,'' \emph{IEEE Intelligent Transportation Systems Magazine},
  pp. 2--19, 2022.

\bibitem{polack2017kinematic}
P.~Polack, F.~Altch{\'e}, B.~d'Andr{\'e}a Novel, and A.~de~La~Fortelle, ``The
  kinematic bicycle model: A consistent model for planning feasible
  trajectories for autonomous vehicles?'' in \emph{2017 IEEE intelligent
  vehicles symposium (IV)}.\hskip 1em plus 0.5em minus 0.4em\relax IEEE, 2017,
  pp. 812--818.

\end{thebibliography}
\bibliographystyle{IEEEtran}



\end{document}